\documentclass{article} %
\usepackage[final]{colm2026_conference}

\usepackage{microtype}
\usepackage{hyperref}
\usepackage{url}
\usepackage{booktabs}
\usepackage{wrapfig}

\usepackage{lineno}
\usepackage{xcolor}
\usepackage{amsmath, amssymb, tikz-cd}
\usepackage[most]{tcolorbox}
\definecolor{darkblue}{rgb}{0, 0, 0.5}
\hypersetup{colorlinks=true, citecolor=darkblue, linkcolor=darkblue, urlcolor=darkblue}

\newcommand{\po}{\mathbb{P}}
\newcommand{\vc}{\mathcal{V}}
\newcommand{\wc}{\mathcal{W}}
\DeclareMathOperator{\rk}{rk}
\newcounter{promptbox}

\usepackage{xspace}
\newcommand{\LiveMathematicianBench}{\textsc{LiveMathematicianBench}\xspace}

\title{{\LiveMathematicianBench}: A Live Benchmark for Research-Level Mathematical Reasoning with Proof Sketches}

\author{Linyang He$^{1,*}$\qquad Qiyao Yu$^{1,*}$\qquad Hanze Dong$^{2,*}$\qquad Baohao Liao$^{2,3}$\\
\textbf{Xinxing Xu$^{2}$\qquad Micah Goldblum$^{1,\dagger}$ \qquad Jiang Bian$^{2,\dagger}$ \qquad Nima Mesgarani$^{1,\dagger}$}\\
\\
$^{1}$Columbia University \qquad $^{2}$Microsoft Research \qquad $^{3}$University of Amsterdam\\
$^{*}$Equal contribution. \qquad $^{\dagger}$Senior authors.\\
\texttt{linyang.he@columbia.edu \quad qy2266@columbia.edu \quad hanzedong@microsoft.com} \\
\\
For the latest benchmark updates, visit \url{https://LiveMathematicianBench.github.io/}
}

\begin{document}

\ifcolmsubmission
\linenumbers
\fi

\maketitle

\begin{abstract}
Mathematical reasoning is widely regarded as a hallmark of human intelligence, and determining whether large language models (LLMs) can meaningfully perform it remains a central question in artificial intelligence and cognitive science. As LLMs are increasingly integrated into scientific workflows, rigorous evaluation of their mathematical capabilities becomes a practical necessity.
Existing mathematical reasoning benchmarks, however, are often limited by synthetic problem settings and growing contamination from widely circulated datasets. 

We present \textbf{\LiveMathematicianBench}, a dynamic multiple-choice benchmark for evaluating research-level mathematical reasoning using recent arXiv papers published after model training cutoffs. By grounding evaluation in newly published theorem statements, the benchmark provides a more realistic testbed for assessing whether models can reason about natural mathematical claims beyond memorized benchmark patterns. \LiveMathematicianBench introduces a taxonomy of thirteen problem categories based on the logical form of theorem statements, enabling fine-grained evaluation across reasoning types such as implication, equivalence, existence, and uniqueness. It further introduces a proof-sketch-guided distractor generation pipeline, in which proof sketches are used to construct plausible but invalid answer choices that reflect misleading proof directions. This design makes the benchmark more sensitive to genuine mathematical understanding rather than surface-level answer matching. We additionally introduce a substitution-resistant evaluation mechanism designed to distinguish answer recognition from substantive mathematical reasoning. 

Evaluation of frontier models shows that the benchmark is far from saturated: the best-performing model, Gemini-3.1-pro-preview, achieves only 43.5\% accuracy in the standard setting. Under substitution-resistant evaluation, accuracy drops sharply across all models: GPT-5.4 achieves the highest score at 30.6\%, while Gemini-3.1-pro-preview falls to 17.6\%, below the 20\% random baseline. A dual-mode protocol comparing performance with and without proof-sketch access reveals that sketches yield consistent gains (e.g., +13.6 pp for Gemini-3.1-pro-preview), suggesting that models can leverage high-level proof strategies to improve reasoning. 

Overall, \LiveMathematicianBench offers a scalable and contamination-resistant testbed for studying research-level mathematical reasoning in large language models.

\end{abstract}

\begin{figure*}[h]
\centering
\includegraphics[width=0.7\textwidth]{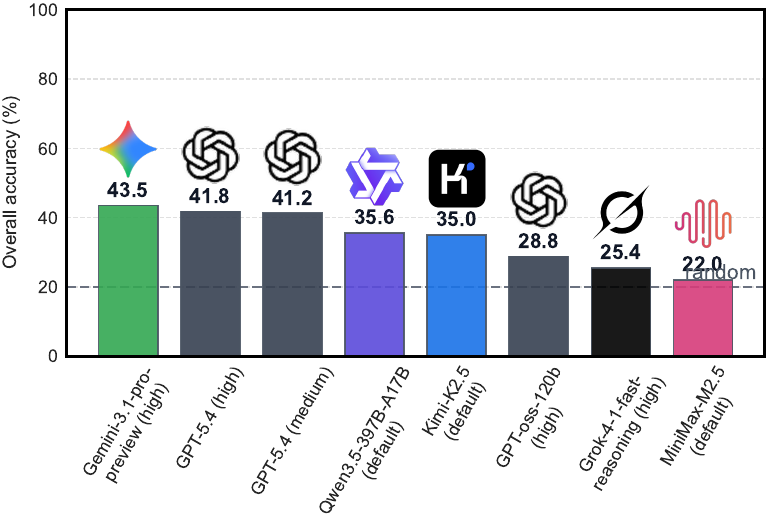}
\caption{Overall accuracy on \LiveMathematicianBench across different LLMs. Even recent frontier models remain far from saturation. Random: 20\%.}
\label{fig:overall_accuracy_logos}
\end{figure*}

\section{Introduction}
As large language models (LLMs) become increasingly integrated into scientific workflows, evaluating their reasoning ability has become a central challenge. Mathematics is a natural domain for this evaluation because it combines precise logical structure with verifiable ground truth. Yet existing benchmarks, including GSM8K \citep{cobbe2021gsm8k}, MATH \citep{hendrycks2021measuring}, AIME, MathVista \citep{lu2023mathvista}, Omni-Math \citep{gao2024omni}, and OlympiadBench \citep{he2024olympiadbench}, are largely synthetic, competition-based, or historically sourced, and are increasingly susceptible to contamination. These limitations reduce their value for assessing research-level mathematical reasoning in frontier models.

The core limitation lies in the fundamental mismatch between “Olympiad-style” competition mathematics and authentic mathematical research. Existing benchmarks predominantly feature calculation-heavy problems that reward heuristic pattern matching and the use of specialized tricks to obtain a closed-form answer. Although challenging, such tasks do not reflect the true nature of mathematics research. 

By \emph{research-level mathematics}, we mean open-ended, non-routine mathematical reasoning that is characterized by abstraction, sensitivity to technical hypotheses, and the lack of a pre-specified solution strategy. Such work
often requires one to reason through layers of definitions, infer subtle logical dependencies, and identify the right conceptual framework in which a problem can even be meaningfully posed. A central difficulty of research lies not only in proving statements, but also in determining what the right question is to ask and what potential statement might be true in the first place. Unlike competition problems, naturally arising research questions do not come with a contrived setup, an intended solution path, or even a clear indication of which tools are relevant. Progress therefore often requires exploration under substantial uncertainty, familiarity with prior literature, and occasional conceptual innovations, which are sometimes sparked by ideas from seemingly unrelated fields of mathematics.

This gap is visible even at the level of theorem comprehension. A model that can solve routine calculus or algebra problems may still fail to track the precise hypotheses of a modularity lifting theorem, a stability result in geometric invariant theory, or a compactness theorem in nonlinear PDE, where small changes in local conditions, regularity assumptions, or finiteness hypotheses can fundamentally alter the statement.
Therefore, to truly assess an LLM's potential as a scientific research assistant, evaluation must shift from measuring the ability to "solve exams" toward measuring the ability to comprehend, interpret, and reason about mathematical hypotheses and statements.

This conceptual mismatch is compounded by a second problem: data contamination. State-of-the-Art (SOTA) models have reached saturation on standard benchmarks, often exceeding 90\% accuracy, yet this performance is brittle. Studies suggest that many problems have leaked into pre-training corpora, allowing models to memorize solution templates rather than derive answers from first principles \citep{glazer2024frontiermath}. Consequently, performance drops precipitously when questions are rephrased or numerical values altered. This reliance on static, potentially contaminated datasets necessitates a shift toward continuous and refreshable evaluations grounded in authentic research artifacts.

Notably, \citet{zhang2025realmath} introduced RealMath, a benchmark derived directly from arXiv papers to reduce contamination risk. 
While RealMath reduces contamination risk and demonstrates that LLMs can engage with research-derived material, its construction prioritizes theorem types that admit straightforward automated verification. As a result, it covers only a limited subset of research-level mathematical reasoning, with less support for tasks involving inequalities, asymptotic relations, or non-constructive arguments, excluding common research tasks such as establishing asymptotic bounds in analysis, deriving classification results in algebraic geometry, or determining modes of convergence in probability. %
These limitations suggest that the central challenge is not only to control contamination risk in benchmarks, but to do so without collapsing research mathematics into a narrowly verifiable subset of problems of limited width and depth.

To address these limitations, we introduce \LiveMathematicianBench, a benchmark framework that expands evaluation along two dimensions: %
the diversity of mathematical reasoning it captures and the use of high-level proof strategies, enabling evaluation beyond statement-level recognition to strategy-level mathematical reasoning.

For the first dimension, we introduce a taxonomy of thirteen distinct problem types defined by the underlying logical forms and tailor question design to each type.
Unlike RealMath, whose questions are restricted to settings with a single uniquely verifiable answer, this framework enables evaluation of mathematical skills that are central to research practice but often considered too ambiguous for automated metrics. These include \textbf{Logical Structure} (distinguishing necessary vs. sufficient conditions via Equivalence and Implication), \textbf{Qualitative Analysis} (establishing Inequalities and Asymptotic bounds), and \textbf{Generalization} (validating Universal quantifiers and structural Bijections).

For the second dimension, we incorporate proof sketches into the benchmark. A proof sketch is a high-level outline of the main ideas and strategic steps of a proof, omitting most technical details. Access to this proof structure makes question construction more targeted and more diagnostic: instead of merely testing whether a model can retrieve a theorem statement, it allows us to design questions about the same result that probe deeper structural understanding. Without such proof-aware design, questions are more likely to be too easy or to reward memorization and surface-level pattern matching. Proof sketches also enable a dual-mode evaluation protocol, where models are assessed both with and without access to high-level proof guidance. This makes it possible to measure whether strategic hints improve mathematical reasoning, and how effectively current models can use them.

Building on these two design principles, we introduce the \textbf{\LiveMathematicianBench}, a dynamic benchmark designed to assess LLMs on research-level mathematical reasoning. Each item is presented as a multiple-choice question (\emph{MCQ}) with five answer options, exactly one of which is correct. Our contributions fall into three categories: benchmark construction, methodological novelty, and scientific questions enabled by the benchmark.

Our contributions are threefold. %
\textbf{1)} We present a contamination-resistant benchmark for research-level mathematical reasoning, built from dynamically sourced post-cutoff arXiv theorems, organized by a logic-based taxonomy, and augmented with curated proof sketches. \textbf{2)} We introduce a proof-aware benchmark construction methodology, including category-specific question design, sketch-adversarial distractor generation, and dual-mode evaluation with and without sketch access. \textbf{3)} We establish a new testbed for studying higher-level mathematical reasoning in LLMs, including logical-form-specific failure modes and the use of proof-level guidance.

\section{Methods}
Our benchmark construction pipeline transforms raw arXiv papers into calibrated research-level multiple-choice questions (MCQs). The pipeline contains seven stages: (1) paper retrieval, (2) \LaTeX\ source extraction, (3) theorem extraction and classification, (4) MCQ generation, (5) stem-only triviality filtering, (6) hardness calibration, and (7) model evaluation. Each stage is designed to preserve mathematical fidelity while increasing question discriminativeness and benchmark difficulty. Human quality control combines a complete review of the 177 final benchmark items with stratified audits of intermediate pipeline outputs. We additionally conduct a small source-author validation study; see Appendix~\ref{app:validation} for details.

\begin{figure*}[t]
\centering
\includegraphics[width=1.0\textwidth]{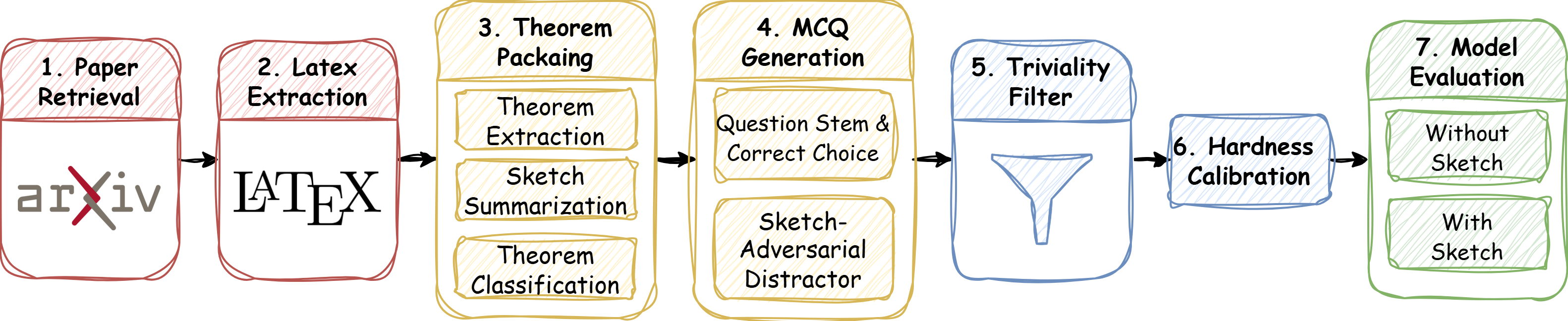}
\caption{Our benchmark construction includes 7 stages.}
\label{fig:pipeline}
\end{figure*}

\noindent\textbf{Stage~1: Paper Retrieval.}%
We query the arXiv API for papers submitted within a target month under the \texttt{math.*} category. The retrieval window is incrementally widened until a configurable maximum number of papers is collected. For each paper, we record its arXiv identifier, submission link, source link, and title. By restricting retrieval to papers published strictly after a model's training cutoff, we ensure that the resulting benchmark is contamination-free by construction.

\noindent\textbf{Stage~2: \LaTeX\ Source Extraction.}
For each retrieved paper, we download its e-print source archive and extract the constituent \texttt{.tex} files. The extractor resolves \verb|\input| and \verb|\include| directives to reconstruct the complete document in its logical order, strips \LaTeX\ comments and comment environments, and concatenates the result into a single full-text representation. This produces a clean, self-contained \LaTeX\ corpus suitable for downstream parsing.

\label{sec:methods}

\noindent\textbf{Stage~3: Theorem Extraction and Classification.}%

\emph{Stage~3a: Hybrid Agentic Extraction.}
Extracting well-formed theorem statements from raw \LaTeX\ is nontrivial due to the heterogeneity of document styles, custom macros, and cross-referencing conventions. We adopt a \emph{hybrid agentic} extraction architecture that balances throughput with robustness:

\begin{enumerate}
    \item \textbf{Rule-based fast path.} A rule-based extractor restricts its search to the \emph{Introduction} section, exploiting the convention that a paper's main theorem is formally stated there. Within the introduction, it identifies all theorem-like environments declared via \verb|\newtheorem| in the preamble (excluding \texttt{remark}-type environments) and applies a \emph{priority ranking}: the standard \texttt{theorem} environment is selected first; only if no \texttt{theorem} block is found does the extractor fall back to custom environments. This design ensures that auxiliary results such as lemmas, propositions, or corollaries, which may appear elsewhere in the introduction, are not mistakenly extracted in place of the main theorem.
    \item \textbf{Agentic fallback.} When the rule-based path fails (e.g., due to an absent Introduction heading or unconventional formatting), an LLM-based agentic extractor is invoked over a larger document window. The extraction prompt explicitly instructs the model to return at most \emph{one} primary theorem-level claim and to prefer \texttt{theorem} environments over \texttt{lemma}, \texttt{proposition}, or \texttt{corollary} when multiple candidates appear, maintaining the same main-theorem priority as the rule-based path.
    \item \textbf{\LaTeX\ normalization.} Custom commands (\verb|\newcommand|, \verb|\def|) are expanded in-line so that downstream modules receive standard mathematical notation.
    \item \textbf{Reference resolution.} Internal cross-references (\verb|\ref|, \verb|\eqref|) are resolved by collecting all labeled environments and substituting their content, producing an \textit{Expanded Theorem} that is semantically self-contained.
    \item \textbf{Context recovery.} A two-layer retrieval mechanism assembles the notational and definitional context needed to make the theorem self-contained. \emph{First}, within the Introduction, paragraph blocks preceding the theorem are scored by (i)~token overlap with the theorem statement, (ii)~the presence of definitional cues (e.g., ``let,'' ``denote,'' ``suppose''), (iii)~mathematical content density, and (iv)~positional proximity, with the two blocks immediately before the theorem always retained. The top-scoring blocks are selected and concatenated in their original order. \emph{Second}, a full-paper pass extracts high-value \emph{anchor terms} from the theorem and scans the entire document for blocks containing definitions, setup, or preliminaries, scoring them by anchor-term hits, section-heading relevance (e.g., \texttt{Setup}, \texttt{Preliminary}), and distance to the theorem. Finally, any content referenced via \verb|\ref| or \verb|\eqref| within the selected blocks is resolved and appended, and the combined context is trimmed to a character budget.
\end{enumerate}

\emph{Stage~3b: Proof-sketch summarization.} A large language model extracts a concise proof sketch for each theorem, capturing the high-level strategy without full formal detail. These sketches serve a dual purpose: they inform adversarial distractor generation (Stage~4) and enable sketch-aware evaluation (Stage~7).
    
\emph{Stage~3c: Logical Taxonomy of Theorems.}
Each extracted theorem is classified into one of \textbf{thirteen} logical categories, listed in Table~\ref{tab:taxonomy}. Classification is performed by a specialized LLM-based classifier. This taxonomy enables fine-grained diagnosis of model reasoning failures across distinct logical forms.

\begin{table}[t]
\centering

\label{tab:taxonomy}
\footnotesize
\begin{tabular}{@{}ll@{}}
\toprule
\textbf{Category} & \textbf{Canonical Form} \\
\midrule
Algorithmic / Constructive  & There exists an algorithm $A$ that computes $f(x)$ \\
Asymptotic / Limit          & As $n \to \infty$, \ldots \\
Biconditional / Equivalence & $A \iff B$ \\
Classification / Bijection  & There is a bijection between $X$ and $Y$ \\
Existence                   & $\exists\, x$ such that $P(x)$ \\
Existential--Universal      & $\exists\, x\; \forall\, y,\; P(x,y)$ \\
Implication                 & $A \implies B$ \\
Independence / Consistency  & $T \not\vdash P$ and $T \not\vdash \neg P$ \\
Inequality / Bound          & $f(x) \leq g(x)$ \\
Nonexistence                & $\not\exists\, x$ such that $P(x)$ \\
Uniqueness                  & $\exists!\, x$ such that $P(x)$ \\
Universal                   & $\forall\, x,\; P(x)$ \\
Universal--Existential      & $\forall\, x\; \exists\, y,\; P(x,y)$ \\
\bottomrule
\end{tabular}
\caption{Logical taxonomy of theorem categories. Each extracted theorem is classified into one of these thirteen types, enabling category-specific question design and evaluation.}
\end{table}

\noindent\textbf{Stage~4: Question--Answer Pair Generation.} 
Stage~4 synthesizes adversarial MCQs via a \textbf{two-stage generative protocol}, each informed by the theorem's logical category and the proof sketch.

\emph{Stage~4a: Question Stem and Correct Choice.}
A key design principle is that every question should read as a \emph{genuine research-type question} about a specific mathematical object or property, rather than a generic request such as ``which is the correct result below.'' To this end, the generator selects one of \textbf{thirteen category-specific system prompts}, one for each logical type in Table~\ref{tab:taxonomy}, each prescribing a tailored stem formulation that exercises the distinguishing logical feature of the category. For example, an equivalence-type theorem is transformed into a question that presents one side of the biconditional and asks ``which statement is equivalent to~$A$?''; an implication-type theorem embeds all hypotheses in the stem and asks ``which is the \emph{strongest} conclusion about \ldots\ that holds under these assumptions?'', etc. With the \emph{strongest} quantifier, we ensure there is a unique answer for questions of all logical types.

Each prompt further instructs the model to (i)~isolate a single \emph{decisive mathematical feature} of the conclusion, such as quantifier scope, sharp bound, exact dependence, existence versus uniqueness, or asymptotic regime, rather than restating the theorem verbatim, and (ii)~define any paper-specific objects or notation inline so that the question is self-contained. Red-flag checks automatically reject stems containing trivializing phrases (e.g., ``which of the following is the strongest result'') or correct options that bundle multiple theorem clauses, triggering a regeneration pass. Finally, a \emph{repair pass} re-examines the draft stem and, operating in \texttt{theorem\_only\_repair} mode, injects any missing definitions or notation from the recovered paper context (Stage~3, Step~5) to guarantee self-containedness without leaking answer-relevant information.

\emph{Stage~4b: Sketch-Adversarial Distractor Generation.}
The model receives the theorem, proof sketch, question stem, and correct choice, and generates \textbf{four distractors} (Options~B--E). The proof sketch plays a critical role: by exposing the ``load-bearing'' logical steps of the argument (e.g., the dependence of constants on hidden parameters, a critical case split, or a uniformity condition that distinguishes effective from non-effective bounds), the generator is guided to engineer distractors that exploit specific mathematical constraints or misconceptions. Each distractor is constructed by one of four general perturbation strategies:

\begin{itemize}
    \item \textbf{Controlled perturbation.} A small surface-level modification to the theorem's conclusion, such as an altered constant, exponent, inequality direction, or parameter range.
    \item \textbf{Semantic weakening.} A statement that is plausible but strictly weaker than the theorem's conclusion, e.g., dropping a uniformity requirement or restricting the domain.
    \item \textbf{Semantic strengthening.} A statement that overstates the conclusion beyond what is proven, e.g., adding uniqueness, higher regularity, or optimality not asserted by the theorem.
    \item \textbf{Property confusion.} A statement concerning a closely related but non-equivalent mathematical object or property, e.g., swapping an $L^2$ norm for an $L^\infty$ norm, or confusing pointwise convergence with uniform convergence.
\end{itemize}

\noindent The concrete realization of each strategy is \emph{category-dependent}: the distractor prompt is tailored to the theorem's logical type (Table~\ref{tab:taxonomy}). For instance, for an equivalence-type theorem the controlled perturbation modifies a quantifier within one side of the biconditional, whereas for a bound-type theorem it alters the exponent or direction of the inequality; for an existence-type theorem the semantic strengthening adds uniqueness not stated in the theorem, whereas for a classification-type theorem it replaces the bijection with a mere surjection. This category-aware design ensures that distractors probe the \emph{decisive logical feature} specific to each theorem type, rather than relying on generic, one-size-fits-all perturbations. An \emph{adversarial revision pass} subsequently re-examines the generated distractors, auditing them for surface-level distinguishability from the correct option and rewriting any that can be rejected without genuine mathematical inspection.

We evaluate the contributions
of sketch access,
category-aware distractor guidance,
and the adversarial revision pass
in Appendix~\ref{app:distractor_ablation}.

\emph{Substitution-Resistant Option Design.}
A common shortcut exploited by strong models is \emph{option substitution}: systematically plugging each candidate answer back into the question stem and verifying consistency, rather than deriving the answer through genuine mathematical reasoning. Process-of-elimination strategies similarly bypass comprehension by narrowing choices through superficial cues. To counteract both shortcuts, we introduce a \emph{substitution-resistant} mechanism: for a configurable fraction of generated items, the correct option is replaced with the meta-option \emph{``One of the remaining options is correct, but a stronger result can be proven.''} Answering correctly now requires the model to (i)~identify the valid option among the distractors and (ii)~independently determine whether a strictly stronger result holds, a task that demands substantive mathematical reasoning about the relative strength of claims and cannot be resolved by mechanical substitution or elimination alone.

\emph{Quality Rubric.}
Each generated MCQ undergoes a quality evaluation scored on a $0$--$8$ rubric comprising four dimensions:
\begin{itemize}
    \item \textbf{Answer Leakage Score (ALS, 0--2):} Measures whether the correct answer is inadvertently revealed by the question stem or distractor phrasing.
    \item \textbf{Tautology Avoidance Score (TAS, 0--2):} Assesses whether the correct option is trivially true or self-evident without mathematical knowledge.
    \item \textbf{Generative Pressure Score (GPS, 0--2):} Evaluates whether distractors exert sufficient ``pressure'' to discriminate genuine understanding from guessing.
    \item \textbf{Distractor Quality Score (DQS, 0--2):} Rates the mathematical plausibility, precision, and diversity of the distractors.
\end{itemize}
MCQs that do not meet a minimum aggregate threshold (5) are discarded.

\noindent\textbf{Stage~5: Rote-Recall Prevented Triviality Filter. }%
Stage~5 ensures that the remaining questions are not trivially solvable from the question stem alone, without reference to the answer choices. A judge model receives \emph{only} the question stem (with all options withheld) and generates a free-form response. A second judge then determines whether this response matches the correct answer (Option~A). Questions for which the stem alone reveals the answer are classified as \emph{stem-trivial} and excluded from the final candidate pool. This filter prevents the benchmark from rewarding rote recall and ensures that genuine option-level reasoning is required.

\noindent\textbf{Stage~6: Hardness Calibration. }%
Stage~6 ensures that the final benchmark is genuinely difficult for state-of-the-art models through a three-step calibration pipeline:

\emph{Step~1: Overgenerate Hard Pool.}
For each stem-nontrivial item, we retain both the original distractor set and one regenerated alternative set, thereby doubling the candidate pool per source theorem. This overgeneration increases the likelihood of retaining at least one maximally challenging variant.

\emph{Step~2: First-Pass Accuracy Test.}
A frontier model is evaluated on the entire candidate pool at moderate reasoning effort. Items that the model answers correctly are flagged as potentially too easy.

\emph{Step~3: Source-Level Hardest Selection.}
For each source theorem, the pipeline compares all surviving candidate variants and selects the \textbf{hardest}, defined as the variant the calibration model answered incorrectly while achieving the highest quality rubric score. Source groups in which all candidates were solved are dropped entirely. This procedure yields the final benchmark subset.

\noindent\textbf{Stage~7: Evaluation Protocol.} We propose a dual-mode evaluation protocol: \textbf{Mode~1: Selection.} The model receives the question stem and all five options, then selects the answer that it judges to be the correct theorem statement. This mode evaluates the model's baseline mathematical comprehension. \textbf{Mode~2: Sketch-Aware Selection.} The model additionally receives the proof sketch as a hint alongside the stem and options. By comparing performance across modes, we isolate the model's ability to synthesize and apply high-level proof strategies, a proxy for \emph{mathematical intuition}. 

To ensure a fair rendering of notations, all questions undergo \LaTeX\ validation via compilation and, when necessary, automated \LaTeX\ repair before presentation to the model.

\section{Benchmark}
\begin{figure*}[t]
\centering
\includegraphics[width=1.0\textwidth]{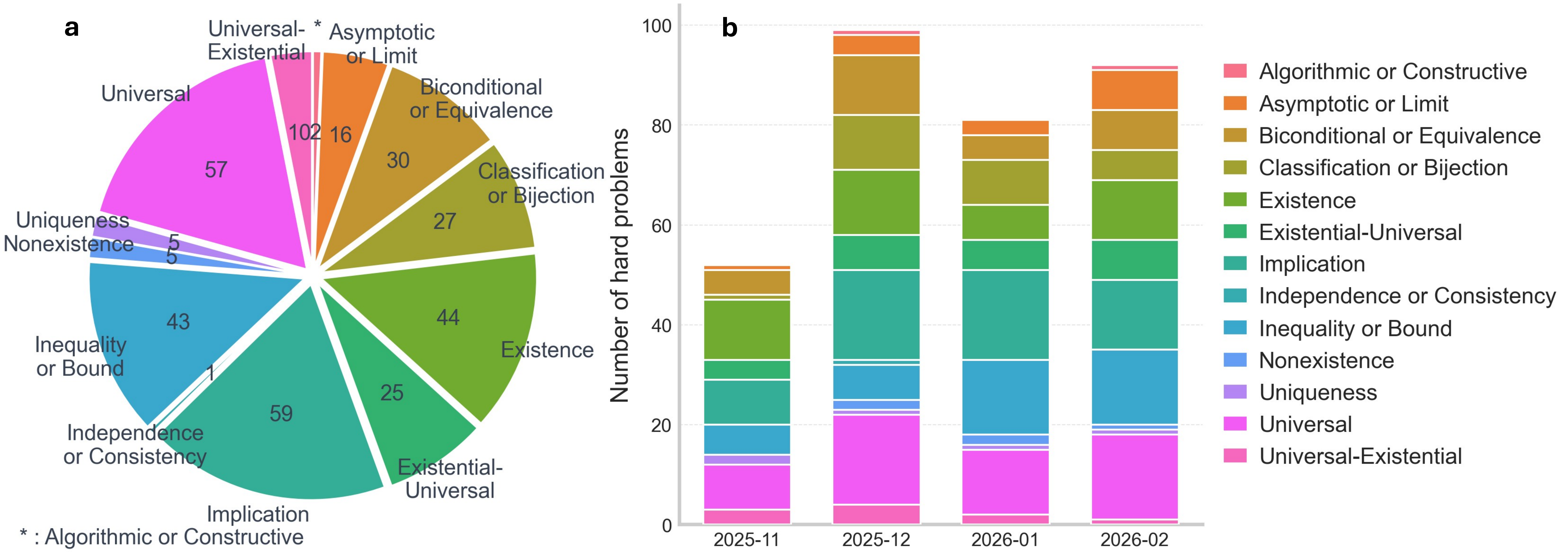}
\caption{Category statistics for the hard split of \LiveMathematicianBench. \textbf{(a)} aggregate distribution of logical categories across all months. \textbf{(b)} month-by-month category breakdown from November 2025 to February 2026. 
Counts correspond to category memberships rather than unique questions, since a single theorem can receive multiple logical labels.}
\label{fig:category_statistics}
\end{figure*}
We summarize the logical-form composition of the hard benchmark split in Figure~\ref{fig:category_statistics}. The hard split contains 177 theorems in total, and a single theorem may belong to multiple logical categories. The benchmark therefore covers a broad range of theorem structures rather than concentrating on a single template: implication and universal statements are the most common categories, followed by existence and inequality/bound problems. The month-level breakdown further shows that benchmark composition changes over time while preserving category diversity, with all slices drawn from papers released after the evaluated models' knowledge cutoffs. Examples can be found in the Appendix \ref{app:taxonomy}.

\section{Evaluation Results}
\label{sec:results}
\begin{figure*}[t]
\centering
\includegraphics[width=1.0\textwidth]{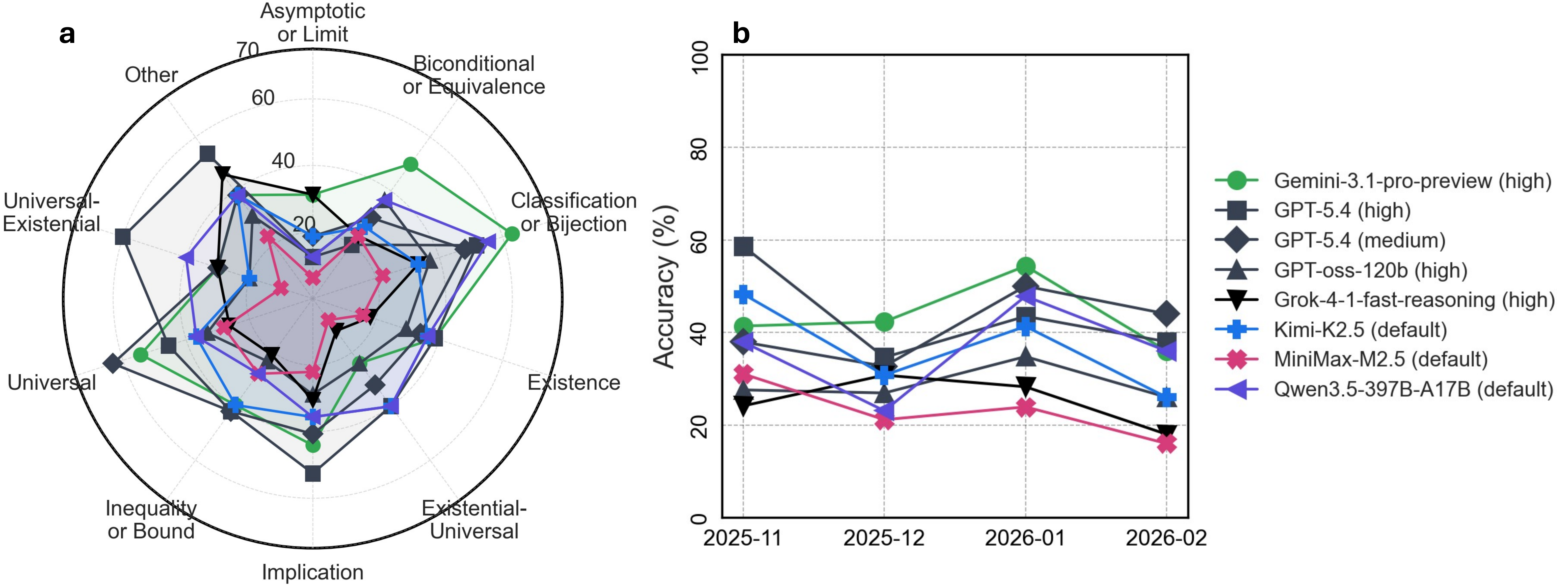}
\caption{
\textbf{(a)} accuracy by logical category aggregated across benchmark months; rare categories with fewer than ten examples are grouped into \emph{Other}. \textbf{(b)} monthly overall accuracy from November~2025 to February~2026. 
}
\label{fig:results_overview}
\end{figure*}

\paragraph{Even frontier models suffer.} Figure~\ref{fig:overall_accuracy_logos} shows overall accuracy on \LiveMathematicianBench. The strongest performance is achieved by Gemini-3.1-pro-preview at 43.5\% overall, followed by GPT-5.4 (high) at 41.8\%. Even the strongest systems remain far from saturation.
\paragraph{Category-wise pattern and monthly trajectories.} Figure~\ref{fig:results_overview}-(a) shows that no single model dominates every reasoning type. Gemini-3.1-pro-preview achieves the strongest overall profile and is particularly competitive on biconditional/equivalence and classification/bijection problems, whereas GPT-5.4 variants are stronger on implication, universal, and inequality/bound items. These categories combined constitute a substantial portion of the benchmark (Figure \ref{fig:category_statistics}). This heterogeneity suggests that research-level mathematical reasoning is not a one-dimensional capability: models with similar aggregate performance can still exhibit sharp weaknesses on particular logical forms. The monthly trajectories in Figure~\ref{fig:results_overview}-(b) further show that performance is unstable across benchmark slices. Several models do best on the 2026/01 subset but perform worse again on the 2026/02 subset.

\begin{figure*}[t]
\centering
\includegraphics[width=1.0\textwidth]{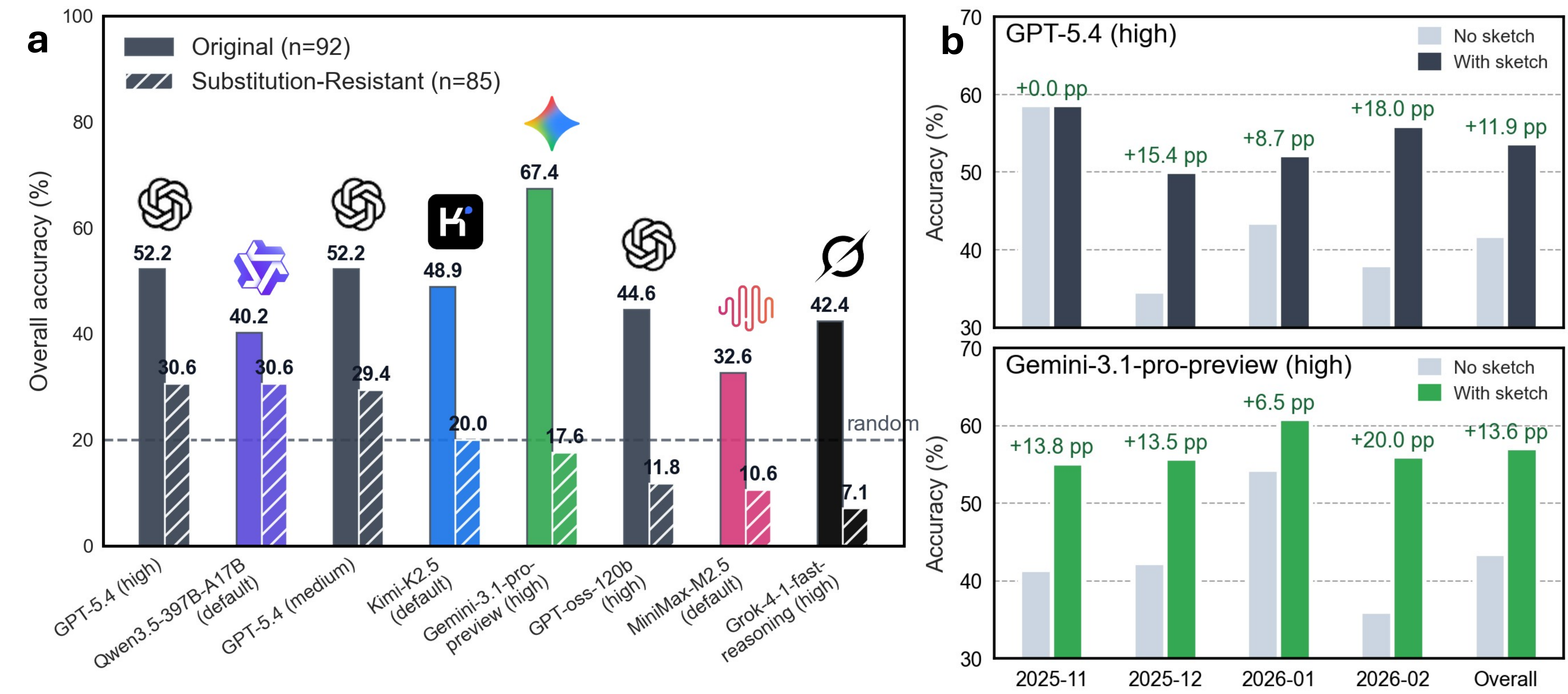}
\caption{\textbf{(a)} Accuracy by answer-choice format on \LiveMathematicianBench. Each model is evaluated separately on the original-choice subset and the substitution-resistant subset. Ranked by substitution-resistant performance.
\textbf{(b)} Accuracy gain from adding proof sketches. Each panel compares the same model setting with and without sketch access across benchmark months. 
}
\label{fig:choice_style_accuracy}
\end{figure*}

\paragraph{Substitution-resistant options sharply reduce accuracy.} Figure~\ref{fig:choice_style_accuracy}-(a) compares model accuracy on the original-choice subset and the substitution-resistant subset, where the correct answer is replaced by a stronger meta-option intended to frustrate shortcut strategies such as direct substitution and surface-level elimination. Across all evaluated systems, the substitution-resistant setting is substantially harder than the original-choice setting. The strongest original-choice performance is achieved by Gemini-3.1-pro-preview at 67.4\%, but its accuracy drops to only 17.6\% on substitution-resistant items, below the 20\% random-guess baseline. 
GPT-5.4 provides a useful contrast: although it ranks second overall, it reaches 52.2\% on original items under both medium and high reasoning effort and still retains 29.4\%--30.6\% on the substitution-resistant subset. GPT-5.4 is therefore more stable under the harder choice design, remaining above the random-guess baseline even when the benchmark removes the direct theorem statement as an option, whereas most other evaluated models fall to chance level or below. More broadly, the remaining difficulty in \LiveMathematicianBench lies not only in understanding the theorem itself, but also in reasoning among nearby mathematical claims once surface matching and answer substitution become ineffective.

\paragraph{Proof sketches yields a consistent improvement.} As shown in Figure~\ref{fig:choice_style_accuracy}-(b), GPT-5.4 (high) improves from 41.8\% to 53.7\% overall, a gain of 11.9\% points, while Gemini-3.1-pro-preview (high) rises from 43.5\% to 57.1\%, a gain of 13.6\% points. 
This pattern suggests that proof sketches are not merely redundant restatements of the theorem. Instead, they provide useful strategic information that helps models disambiguate closely related answer choices and identify the load-bearing structure of the argument.

\section{Endnote and Discussion}

From aggregate accuracy alone, Gemini-3.1-pro-preview (high) achieves the highest score, with GPT-5.4 variants close behind by less than 3\%. However, a category-level analysis in Section~\ref{sec:results} (Fig.~\ref{fig:results_overview}(a)) reveals additional structure that supports the first central design choice of \LiveMathematicianBench: organizing problems by logical type. Performance varies meaningfully across logical forms in ways that are not visible from aggregate scores alone. This suggests that logical structure may be an important axis along which current models differ, and it raises a broader interpretability question: why are some models systematically stronger on certain forms of reasoning than others? If such patterns persist at scale, they may also have practical implications for mathematical workflows, since different models may prove more useful for different kinds of reasoning tasks. In this sense, a logic-based taxonomy complements traditional field-based categorization by capturing structural similarities that cut across mathematical areas. More broadly, this suggests that \LiveMathematicianBench\ can serve not only as a leaderboard benchmark, but also as a diagnostic testbed for studying structure-specific strengths and weaknesses in mathematical reasoning.
This diagnostic value is especially relevant because the benchmark is constructed from recent post-cutoff arXiv papers, making it less vulnerable to rapid saturation and contamination than static mathematics benchmarks. In this sense, \LiveMathematicianBench\ contributes not only a new dataset, but also a more durable evaluation setting for tracking progress in research-level mathematical reasoning.

Substitution-resistant options (Stage 4b) are designed to test whether a model can recognize that the \emph{semantically weakened} option is true while also inferring that a strictly stronger statement follows from the hypotheses. Success on such questions therefore requires an additional deductive step. In this sense, substitution-resistant questions provide a useful probe of whether a model can move beyond recognition toward a limited form of mathematical conjecturing. This is precisely the kind of ability mathematicians hope AI systems may leverage on when facilitating research, as illustrated by the work of \cite{georgiev2025mathematicalexplorationdiscoveryscale} using AlphaEvolve for mathematical exploration and discovery. As highlighted in Section~\ref{sec:results}, all tested models perform substantially worse with substitution-resistant options (Fig.~\ref{fig:choice_style_accuracy}{(a)}). This gap suggests that current models rely heavily on option substitution in standard multiple-choice settings. More importantly, it underscores the importance of adversarial option design for isolating genuine reasoning ability. At its core, it shows that benchmark design can materially affect the capabilities we appear to measure. From considerations of both overall benchmark performance and performance under substitution-resistant evaluation, GPT-5.4 (high) and Gemini-3.1-pro-preview (high) emerge as the two strongest models on \LiveMathematicianBench.

Finally, we have observed earlier that providing proof sketches at inference time induces significant performance improvements in the two highest-scoring models (Fig.~\ref{fig:choice_style_accuracy}{(b)}). Since proof sketches are extracted from the source papers and typically summarize the central ideas of the proof at a high level, this result suggests that models can often use strategic mathematical guidance without full derivational detail, demonstrating some similarities to how expert researchers work. These findings further support the second design philosophy of \LiveMathematicianBench: incorporating proof strategy into evaluation rather than restricting assessment to statement-level recognition alone. More broadly, they indicate that the benchmark can distinguish not only whether a model reaches the correct answer, but also how its performance changes when supplied with information closer to the level of mathematical strategy used by human researchers. Collectively, these features make \LiveMathematicianBench\ useful not only for comparing current models, but also for future work on proof-guided reasoning, tool use, and interactive mathematical exploration.

\bibliography{example_paper}
\bibliographystyle{colm2026_conference}

\appendix
\onecolumn
\section{Related Work}

\paragraph{Standard and Advanced Mathematical Benchmarks}
The evaluation of mathematical reasoning in large language models has traditionally relied on static datasets spanning elementary to undergraduate-level problems. Early benchmarks such as GSM8K \citep{cobbe2021gsm8k} and MATH \citep{hendrycks2021measuring} helped establish common evaluation standards for quantitative and symbolic reasoning. As model performance has improved, however, these benchmarks have become less discriminative for frontier systems.

Subsequent work has therefore shifted toward more challenging settings. The Advanced Reasoning Benchmark (ARB) \citep{sawada2023arb} and GHOSTS \citep{frieder2023mathematical} extend evaluation toward graduate-level mathematics, with GHOSTS further emphasizing natural language interaction and expert assessment. Other efforts focus on competition-level mathematics, including OlympiadBench \citep{he2024olympiadbench}, PutnamBench \citep{tsoukalas2024putnambench}, and Omni-Math \citep{gao2024omni}. These benchmarks substantially increase difficulty, but because they are derived from historical competition archives, they may be more susceptible to contamination from pretraining data and benchmark-specific memorization.

\paragraph{Contamination Mitigation, Formal Mathematics, and Frontier-Level Evaluation}
Several recent benchmarks attempt to mitigate contamination more directly. Putnam-AXIOM \citep{gulati2025putnam} perturbs existing problems through programmatic modifications such as renaming variables or altering constants. While this strategy can reduce exact overlap with known solutions, it preserves much of the original problem structure. The AI Mathematical Olympiad (AIMO) \citep{xtx2024aimo} instead introduces newly created olympiad-style problems, and MiniF2F \citep{zheng2021minif2f} focuses on formally verified mathematics across theorem proving systems such as Lean and Isabelle. These directions address important aspects of evaluation reliability, but they remain centered on competition-style or formalized problem settings.

Moving beyond competition settings, First Proof (\cite{abouzaid2026proof}) curates a collection of ten original, research-level problems drawn directly from the authors' active work. While this approach effectively guarantees the absence of data contamination, the extremely small sample size limits its utility as a comprehensive evaluation benchmark. 

FrontierMath \citep{glazer2024frontiermath} is more closely related to our setting in that it targets highly difficult mathematics under reduced contamination risk through a larger set of expert-authored problems. However, similar to First Proof,  tailored problem authoring requires substantial dedicated human effort, which inherently limits scalability.

\LiveMathematicianBench differs in both construction and evaluation emphasis. Rather than relying on manually authored problems, it draws from a continuously growing stream of recent arXiv papers, enabling a scalable and refreshable benchmark construction pipeline. In addition, whereas prior benchmarks largely emphasize answer derivation, our framework explicitly evaluates structural comprehension through theorem-level logical categories and proof-sketch-guided distractors that probe sensitivity to plausible but invalid reasoning directions.
\section{Human Validation Protocol}
\label{app:validation}

All prompts used in theorem extraction, classification, question generation, and filtering were authored by a research-level mathematician. For quality control, the final released set of 177 benchmark items was reviewed for theorem fidelity, self-containedness, answer uniqueness, and distractor quality. To audit upstream pipeline quality, we additionally performed stratified reviews of intermediate outputs, sampling approximately 10 cases per logical category for theorem extraction, category assignment, proof-sketch quality, and MCQ generation. Given the unusual breadth and depth of the mathematical content covered by the pipeline, these reviews were intended as benchmark quality control, assessing whether outputs were mathematically sound, internally consistent, and benchmark-ready, rather than as full peer-review-level sense of independent verification of every source-paper claim.

\subsection{Source-Author Validation}
\label{app:author_validation}

To complement the internal review,
we conducted a validation study
with authors of the source papers.

For each item included in the study,
we sent the source-paper authors
the extracted theorem,
question stem,
proof sketch,
and multiple-choice options.

We asked them to assess:

(i) whether the item provided
sufficient definitions and context;

(ii) whether it captured
the theorem's load-bearing
mathematical content; and

(iii) whether the question stem
and selected answer
accurately rephrased
the original theorem statement.

Across 10 valid responses,
the mean scores were
7.8 out of 10
for context sufficiency,
7.4 out of 10
for capturing the theorem's
load-bearing content,
and 7.3 out of 10
for rephrasing accuracy.

These results provide
preliminary expert evidence
that the generated items
are aligned with
their source theorems.

The study is limited
by its small sample size
and does not constitute
independent verification
of every source-paper theorem
or every released benchmark item.

\paragraph{Scope of Correctness Validation.}

\LiveMathematicianBench\
evaluates whether generated items
faithfully represent claims stated
in recent source papers.

It does not independently certify
that every theorem in those papers
is mathematically correct.

Requiring peer-review-level verification
of every source claim
would substantially delay
benchmark construction
and weaken its freshness
and contamination resistance.

The pipeline mitigates this risk
by selecting theorem statements
with accompanying proof context,
reviewing final items
for fidelity and answer uniqueness,
and obtaining feedback
from source-paper authors.

These safeguards reduce,
but do not eliminate,
errors introduced during extraction
or inherited from the source papers.

We now provide an example demonstrating how the combination of prompt design and human validation ensures that all distractors are demonstrably false, thereby guaranteeing the overall validity of the MCQs.

In Appendix \ref{app:taxonomy}, Example \ref{box:exiuni} is generated from an Existential-Universal type theorem, and the correct option generated from the theorem is replaced by the correct substitution-resistant option. When generating distractors, the system prompt (see Appendix \ref{app:prompt}, Stage 4b) forces the LLM to include a meta description for each distractor indicating the type of perturbation strategies used in its construction. For Example \ref{box:exiuni}, the meta description reads:

\begin{verbatim}
     "meta": {
          "weaker_true_label": "C",
          "false_labels": [
            "B",
            "D",
            "E"
          ],
          "wildcard_false_label": "B"
        },
        "sketch_usage_meta": [
          {
            "label": "B",
            "sketch_hook_type": "counting_estimate",
            "tampered_component": "counts signed cubic discriminants instead of
            cubic fields",
            "template_used": "wildcard"
          },
          {
            "label": "C",
            "sketch_hook_type": "finiteness",
            "tampered_component": "dropped the lower bound guaranteeing 
            at least k/2 cubic fields in the interval",
            "template_used": "weaker_true"
          },
          {
            "label": "D",
            "sketch_hook_type": "counting_estimate",
            "tampered_component": "threshold k/2 replaced by (k+1)/2",
            "template_used": "stronger_trap"
          },
          {
            "label": "E",
            "sketch_hook_type": "regularity",
            "tampered_component": "removed the epsilon-loss in the exponent",
            "template_used": "stronger_trap"
          }
        ]
\end{verbatim}

We first notice that option \textbf{C} is the "weaker\_true\_label", meaning that it is strictly weaker than the theorem's assertion (which is always true if we assume the theorem itself is true). However, since the question asks for the 'strongest' true statement, option \textbf{C} is not correct.

Option \textbf{B} is the "wildcard\_false\_label", meaning that it is designed to be false but with a question specific tempering. It counts the number of integers within that range that are determinants of cubic fields. This is clearly not the same as counting the number of cubic fields, as multiple non-isomorphic fields can share the same discriminant. Therefore, it is not realistic for a smaller set to have the same lower bound. This reasoning aligns with the "tempered\_component" field provided in the meta description. Moreover, this option critically uses the proof sketch.

Option \textbf{D} and \textbf{E} are false labels tempered with "stronger\_trap", meaning that they are strengthened versions of the theorem statement in different ways. Option \textbf{D} increases the lower bound of the set being counted, making the statement stronger. However, the lower bound is obtained by explicit construction and a bigger lower bound cannot be achieved. Option \textbf{E} modifies the count for the specific integers $d$ by removing the epsilon factor in the exponent. The epsilon-dependence in the original theorem comes from deep conjectures in the subject of counting number fields, which cannot be removed.

To conclude, taking the original theorem as the ground truth, we can see from a combination of meta description and human expertise that the distractors are false and the MCQ is valid. In particular, this example demonstrates that our prompt design alone has guaranteed the mathematical falsity of the distractors to a great extent. 

\section{Detailed Logical Taxonomy of Theorem Categories}
\label{app:taxonomy}
This appendix provides extended descriptions of the thirteen logical theorem categories listed in Table~\ref{tab:taxonomy}. The taxonomy focuses exclusively on the \emph{logical form} of a theorem statement, independent of its mathematical topic or context.
Each category is accompanied by its canonical logical form and a representative example.

\paragraph{1. Algorithmic / Constructive.}
\emph{Canonical form:} There exists an algorithm $A$ that computes $f(x)$, or ``object $X$ can be explicitly constructed.''
This category covers theorems providing an explicit or computable method to obtain an object.
\emph{Example:} ``There exists a polynomial-time algorithm that computes a maximum matching in any bipartite graph.''

\begin{tcolorbox}[
    enhanced,
    colback=blue!4,
    colframe=blue!55!black,
    colbacktitle=blue!18,
    coltitle=black,
    boxrule=0.8pt,
    arc=2mm,
    left=6pt,
    right=6pt,
    top=6pt,
    bottom=6pt,
    title=\textbf{{\LiveMathematicianBench} Example},
]
\footnotesize

\textbf{Original Theorem} (\cite{marques2025geometricprogressionsmeetzeckendorf} )i@\par\smallskip
\label{thm:main_window}
Let $u\ge 1$, $q\ge 2$ be integers, and let $\mathcal F$ be a finite family of
non-empty binary words.  Let $L=\max\{|f|:f\in\mathcal F\}$ and fix $M\ge L$.
Then the set
\[
   S_u^{(M)}=\{n\ge 0:\ uq^n\in \mathcal K_{\mathcal F}^{(M)}\}
\]
is either finite or ultimately periodic.  That is, there exist integers $n_0\ge 0$ and $p\ge 1$
such that
\[
   n\in S_u^{(M)}\ \Longleftrightarrow\ n+p\in S_u^{(M)}\qquad(n\ge n_0).
\]
Moreover, one can effectively compute such a pair $(n_0,p)$ and decide whether
$S_u^{(M)}$ is finite or infinite.  In the finite case, the same procedure
outputs all elements of $S_u^{(M)}$.

\vspace{1em}
\hrule
\vspace{1em}

\textbf{Proof Sketch} \par\smallskip
To prove the main theorem, the argument is ``entirely finite-state''. The key input is that ``multiplication by a fixed integer $q$ in the Fibonacci numeration system is realizable by a finite sequential transducer'', and when ``reading digits from the least significant side'' this yields (for each $M$) ``a deterministic update rule on the set of length-$M$ windows $\{0,1,\#\}^M$.'' Composing this with ``the finite automaton that checks avoidance of $\mathcal F$ inside the window'' produces ``a deterministic dynamical system on a finite state space.'' The orbit corresponding to $(uq^n)_{n\ge 0}$ is thus ``eventually periodic'', and ``which yields ultimate periodicity of $S_u^{(M)}$ as an immediate corollary.'' Finally, ``the same finite-state construction gives an explicit algorithm to compute the eventual period'' (hence to compute $(n_0,p)$ and decide finiteness, and list elements in the finite case).

\end{tcolorbox}
\begin{tcolorbox}[
    enhanced,
    colback=green!4,
    colframe=green!55!black,
    colbacktitle=green!18,
    coltitle=black,
    boxrule=0.8pt,
    arc=2mm,
    left=6pt,
    right=6pt,
    top=6pt,
    bottom=6pt,
    title=\textbf{{\LiveMathematicianBench} Example \refstepcounter{promptbox}\thepromptbox~ (Question \& Answers)\label{box:algo}},
]
\footnotesize

\textbf{Generated Question} \par\smallskip
Let $(F_n)_{n\ge 0}$ be the Fibonacci numbers, defined by $F_0=0$, $F_1=1$, and $F_{n+2}=F_{n+1}+F_n$. Every integer $N\ge 1$ has a unique Zeckendorf representation
\[
N=\sum_{i=2}^k \varepsilon_i F_i,
\]
where $\varepsilon_i\in\{0,1\}$, $\varepsilon_k=1$, and $\varepsilon_i\varepsilon_{i+1}=0$. Write $\operatorname{rev}(Z(N))=\varepsilon_2\varepsilon_3\cdots \varepsilon_k$ for the Zeckendorf digits read from least significant to most significant, and let $\widetilde Z(N)=\operatorname{rev}(Z(N))\,\#^{\omega}$, where $\#\notin\{0,1\}$ is a padding symbol.

Let $u\ge 1$ and $q\ge 2$ be integers, let $\mathcal F$ be a finite family of non-empty binary words, let $L=\max\{|f|:f\in\mathcal F\}$, and fix $M\ge L$. Define
\[
\mathcal K_{\mathcal F}^{(M)}=\{N\ge 1:\text{ the prefix of length }M\text{ of }\widetilde Z(N)\text{ contains no factor from }\mathcal F\},
\]
and
\[
S_u^{(M)}=\{n\ge 0: u q^n\in \mathcal K_{\mathcal F}^{(M)}\}.
\]
Which is the strongest conclusion that can be justified about $S_u^{(M)}$?

\vspace{1em}
\hrule
\vspace{1em}

\textbf{Answer Choices} \par\smallskip
\textbf{A.} $S_u^{(M)}$ is either finite or ultimately periodic: there exist integers $n_0\ge 0$ and $p\ge 1$ such that
\[
n\in S_u^{(M)}\iff n+p\in S_u^{(M)}\qquad (n\ge n_0).
\]
Moreover, one can effectively compute such a pair $(n_0,p)$ and decide whether $S_u^{(M)}$ is finite or infinite; in the finite case, the same procedure outputs all elements of $S_u^{(M)}$. \textcolor{green!60!black}{\checkmark}

\textbf{B.} $S_u^{(M)}$ is either finite or ultimately periodic: for each fixed $q$, $\mathcal F$, and $M$, there exist integers $n_0\ge 0$ and $p\ge 1$, independent of $u$, such that for every $u\ge 1$ one has
\[
n\in S_u^{(M)}\iff n+p\in S_u^{(M)}\qquad (n\ge n_0).
\]
Moreover, one can effectively compute such a universal pair $(n_0,p)$ and decide for each $u$ whether $S_u^{(M)}$ is finite or infinite; in the finite case, the same procedure outputs all elements of $S_u^{(M)}$.

\textbf{C.} $S_u^{(M)}$ is ultimately periodic: there exist integers $n_0\ge 0$ and $p\ge 1$ such that
\[
n\in S_u^{(M)}\iff n+p\in S_u^{(M)}\qquad (n\ge n_0).
\]
Moreover, one can effectively compute such a pair $(n_0,p)$.

\textbf{D.} $S_u^{(M)}$ is either finite or ultimately periodic: there exist integers $n_0\ge 0$ and $p\ge 1$ such that
\[
n\in S_u^{(M)}\iff n+p\in S_u^{(M)}\qquad (n\ge n_0).
\]
Moreover, one can always choose these with $n_0<3^M$ and $1\le p\le 3^M$, and effectively compute such a pair together with a decision procedure for whether $S_u^{(M)}$ is finite or infinite; in the finite case, the same procedure outputs all elements of $S_u^{(M)}$.

\textbf{E.} $S_u^{(M)}$ is either finite or ultimately periodic: there exist integers $n_0\ge 0$ and $p\ge 1$ such that
\[
n\in S_u^{(M)}\iff n+p\in S_u^{(M)}\qquad (n\ge n_0).
\]
Moreover, one can effectively decide whether $S_u^{(M)}$ is finite or infinite and, in the infinite case, compute an eventual period $p$; however, in general there is no procedure that always computes a valid preperiod $n_0$ or lists all elements when $S_u^{(M)}$ is finite.

\end{tcolorbox}

\newpage
\paragraph{2. Asymptotic / Limit.}
\emph{Canonical form:} As $n \to \infty$ (or another limiting regime), a quantity behaves as described.
This category covers theorems describing limiting or asymptotic behavior, including exact limits, asymptotic equivalences ($\sim$), and big-$O$/little-$o$ estimates.
\emph{Example:} ``The number of primes less than $x$ satisfies $\pi(x) \sim x / \ln x$ as $x \to \infty$.''

\begin{tcolorbox}[
    enhanced,
    breakable,
    colback=blue!4,
    colframe=blue!55!black,
    colbacktitle=blue!18,
    coltitle=black,
    boxrule=0.8pt,
    arc=2mm,
    left=6pt,
    right=6pt,
    top=6pt,
    bottom=6pt,
    title=\textbf{{\LiveMathematicianBench} Example  (Theorem \& Sketch)},
]
\footnotesize

\textbf{Original Theorem} (\cite{tang2025harmoniclcmpatternssunflowerfree}) \par\smallskip
Fix an integer $k \ge 3$. Then, as $N \to \infty$, 
\[f_k(N) \ge (\log N)^{c_k-o(1)},\] where
\[c_k := \frac{k-2}{e ((k-2)!)^{1/(k-2)}}.\]

\vspace{1em}
\hrule
\vspace{1em}

\textbf{Proof Sketch} \par\smallskip
To prove the main theorem, they use a ``prime bucketing'' construction: partition primes in a suitable interval into many disjoint blocks of comparable harmonic sum, and form squarefree integers by choosing exactly $k-2$ primes from each block. A short combinatorial lemma then forces LCM-$k$-freeness, while the harmonic sum factorizes and can be optimized over a free parameter.

\end{tcolorbox}

\begin{tcolorbox}[
    enhanced,
    breakable,
    colback=green!4,
    colframe=green!55!black,
    colbacktitle=green!18,
    coltitle=black,
    boxrule=0.8pt,
    arc=2mm,
    left=6pt,
    right=6pt,
    top=6pt,
    bottom=6pt,
    title=\textbf{{\LiveMathematicianBench} Example \refstepcounter{promptbox}\thepromptbox~ (Question \& Answers)\label{box:asymptotic} },
]
\footnotesize

\textbf{Generated Question} \par\smallskip
Fix an integer $k \ge 3$. For $N \ge 1$, write $[N] := \{1,2,\dots,N\}$, and define
\[
f_k(N) := \max\left\{\sum_{a\in A}\frac1a: A \subseteq [N] \text{ is LCM-$k$-free}\right\},
\]
where $A$ is called LCM-$k$-free if there do not exist distinct $a_1,\dots,a_k \in A$ such that all pairwise least common multiples are equal, i.e.
\[
\operatorname{lcm}(a_i,a_j) = \operatorname{lcm}(a_1,a_2) \qquad (1 \le i < j \le k).
\]
Define
\[
c_k := \frac{k-2}{e\,((k-2)!)^{1/(k-2)}}.
\]
As $N \to \infty$, which is the strongest asymptotic lower bound that holds for the quantity $f_k(N)$?

\vspace{1em}
\hrule
\vspace{1em}

\textbf{Answer Choices} \par\smallskip
\textbf{A.} One of the remaining options is correct, but a stronger result can be proven. \textcolor{green!60!black}{\checkmark}

\textbf{B.}
\[
f_k(N) \ge (\log N)^{c_k+o(1)} \qquad (N \to \infty),
\]
with $c_k = \dfrac{k-2}{e\,((k-2)!)^{1/(k-2)}}$.

\textbf{C.}
\[
f_k(N) \ge (\log N)^{c_k-o(1)} \qquad (N \to \infty),
\]
for some positive constant $c_k$ depending only on $k$.

\textbf{D.}
\[
f_k(N) \ge (\log N)^{c_k-o(1)}
\]
for every sufficiently large $N$, where now $c_k = \dfrac{k-2}{((k-2)!)^{1/(k-2)}}$.

\textbf{E.}
\[
f_k(N) \ge (\log N)^{\frac{k-1}{e\,((k-1)!)^{1/(k-1)}}-o(1)} \qquad (N \to \infty).
\]

\end{tcolorbox}

\newpage
\paragraph{3. Biconditional / Equivalence.}
\emph{Canonical form:} $A \iff B$, or ``the following are equivalent.''
This category covers theorems asserting that two (or more) conditions are logically equivalent. The theorem may present the equivalence as a single biconditional or as a list of mutually equivalent statements.
\emph{Example:} ``A ring $R$ is Noetherian if and only if every ideal of $R$ is finitely generated.''

\begin{tcolorbox}[
    enhanced,
    colback=blue!4,
    colframe=blue!55!black,
    colbacktitle=blue!18,
    coltitle=black,
    boxrule=0.8pt,
    arc=2mm,
    left=6pt,
    right=6pt,
    top=6pt,
    bottom=6pt,
    title=\textbf{{\LiveMathematicianBench} Example (Theorem \& Sketch)},
]
\footnotesize

\textbf{Original Theorem} (\cite{bi2025quantitativestabilitycliffordtorus}) \par\smallskip
Let $\Sigma\subset\mathbb{S}^3$ be a smooth embedded surface of genus at least $1$.  
Then its canonical five-parameter family 
\[
\{\Sigma_{(v,t)}\}_{(v,t)\in \mathring{B}^4\times[-\pi,\pi]}
\]
satisfies
\[
\max_{(v,t)\in \mathring{B}^4\times[-\pi,\pi]}
\mathrm{Area}\bigl(\Sigma_{(v,t)}\bigr)\;\geqslant\; 2\pi^2.
\]
Moreover, equality holds if and only if $\Sigma$ is the Clifford torus.

\vspace{1em}
\hrule
\vspace{1em}

\textbf{Proof Sketch} \par\smallskip
Using the Marques--Neves five-parameter family $\{\Sigma_{(v,t)}\}$, the key point is to show that if a genus--one surface $\Sigma$ satisfies $\mathcal{W}(\Sigma)\leqslant 2\pi^2+\delta^2$, then after a \emph{conformal} transformation one gets a surface $\Sigma'$ with $\mathrm{Area}(\Sigma')\geqslant 2\pi^2-C\delta^2$.

To prove the main theorem, we use that the full $(v,t)$-family contains a slice of area at least $2\pi^2$, and since $v$ encodes conformal transformations and $t$ is signed distance, the aim is to realize such a high-area slice \emph{within the conformal} four-parameter subfamily; for this one must control $t$ in terms of the energy gap $\delta$, so the maximizing slice lies close to $t=0$.

A closer examination of the Heintze--Karcher inequality used by Marques--Neves shows that, once the family is built from the quantitatively normalized surface, the problem reduces to proving the maximizing parameter $(v_0,t_0)$ stays a definite distance away from the boundary of the conformal group, i.e. there exists $\eta_1>0$ such that $|v_0|\leqslant 1-\eta_1$ (reduced later; technical proof deferred to an appendix).

Geometric intuition for this uniform bound: as $|v|\to1$, $\Sigma_{(v,t)}$ becomes quantitatively close to a geodesic sphere, forcing area strictly below $2\pi^2$, so such $v$ cannot maximize. More precisely, for $t=0$, $\Sigma_v=\Sigma_{(v,0)}$ is Hausdorff close to a geodesic sphere $S$, so $\mathrm{Area}(\Sigma_v)$ is close to $\mathrm{Area}(S)\le 4\pi$, and for small $|t|$ one uses
\[
\bigl|\mathrm{Area}(\Sigma_{(v,t)})-\mathrm{Area}(\Sigma_v)\bigr|\lesssim |t|\,\mathcal{W}(\Sigma_v)
\]
so the area stays $<5\pi$. For larger $|t|$, either (i) if $\Sigma_v$ stays away from the poles, an Euclidean-model argument shows $\Sigma_{(v,t)}$ is a small Lipschitz graph over a geodesic sphere hence has area $<5\pi$, or (ii) if $\Sigma_v$ lies close to a pole, comparison geometry gives $\mathrm{Area}(\Sigma_{(v,t)})<5\pi$. Thus for $|v|$ sufficiently close to $1$, $\mathrm{Area}(\Sigma_{(v,t)})<5\pi<2\pi^2$, proving $|v_0|\le 1-\eta_1$.

With this uniform control of $v$, the remaining task is to derive a quantitative stability estimate for minimal surfaces; the paper then uses the earlier area lower bound to reduce Willmore stability to minimal-surface stability, and later performs linearization/stability analysis to obtain the quantitative bounds (conformal factor, conformal structure, and $W^{2,2}$-distance to the Clifford torus), thereby proving the theorem.

\end{tcolorbox}
\begin{tcolorbox}[
    enhanced,
    colback=green!4,
    colframe=green!55!black,
    colbacktitle=green!18,
    coltitle=black,
    boxrule=0.8pt,
    arc=2mm,
    left=6pt,
    right=6pt,
    top=6pt,
    bottom=6pt,
    title=\textbf{{\LiveMathematicianBench} Example  \refstepcounter{promptbox}\thepromptbox~ (Question \& Answers)\label{box:biconditional}},
]
\footnotesize

\textbf{Generated Question} \par\smallskip
Let $\Sigma\subset \mathbb{S}^3\subset \mathbb{R}^4$ be a smooth embedded surface in the round 3-sphere, with genus at least $1$. For $v\in \mathring{B}^4=\{v\in\mathbb{R}^4:|v|<1\}$, let
\[
F_v(x)=(1-|v|^2)\frac{x-v}{|x-v|^2}-v
\]
be the associated conformal transformation of $\mathbb{S}^3$, and set $\Sigma_v:=F_v(\Sigma)$. If $\mathbb{S}^3\setminus \Sigma=A\cup A^*$ is the decomposition into connected components, let $A_v:=F_v(A)$, let $d_v$ be the signed distance to $\Sigma_v$ (positive on $A_v$, negative on the other component), and define
\[
\Sigma_{(v,t)}=\partial\{x\mid d_v(x)<t\}, \qquad (v,t)\in\mathring{B}^4\times[-\pi,\pi].
\]
Thus $\{\Sigma_{(v,t)}\}_{(v,t)\in \mathring{B}^4\times[-\pi,\pi]}$ is the canonical five-parameter family of $\Sigma$. Which quantitative area estimate holds for this family, and in which case does equality occur?

\vspace{1em}
\hrule
\vspace{1em}

\textbf{Answer Choices} \par\smallskip
\textbf{A.}
\[\max_{(v,t)\in \mathring{B}^4\times[-\pi,\pi]} \operatorname{Area}(\Sigma_{(v,t)})\ge 2\pi^2,\] and equality holds if and only if $\Sigma$ is the Clifford torus. \textcolor{green!60!black}{\checkmark}

\textbf{B.}
\[\max_{(v,t)\in \mathring{B}^4\times[-\pi,\pi]} \operatorname{Area}(\Sigma_{(v,t)})\le 2\pi^2,\] and equality holds if and only if $\Sigma$ is the Clifford torus.

\textbf{C.}
\[\max_{(v,t)\in \mathring{B}^4\times[-\pi,\pi]} \operatorname{Area}(\Sigma_{(v,t)})\ge 2\pi^2.\]

\textbf{D.}
\[\max_{v\in \mathring{B}^4} \operatorname{Area}(\Sigma_v)\ge 2\pi^2,\] and equality holds if and only if $\Sigma$ is the Clifford torus.

\textbf{E.}
\[\max_{(v,t)\in \mathring{B}^4\times[-\pi,\pi]} \operatorname{Area}(\Sigma_{(v,t)})\ge 2\pi^2,\] and equality holds whenever $\Sigma$ is conformally equivalent to the Clifford torus.

\end{tcolorbox}

\vspace{1em}

\newpage
\paragraph{4. Classification / Bijection.}
\emph{Canonical form:} There is a bijection between $X$ and $Y$, or ``objects of type $A$ are completely classified by objects of type $B$.''
This category covers theorems establishing a one-to-one correspondence or a complete classification between two classes of mathematical objects.
\emph{Example:} ``Finite simple groups are classified into cyclic groups of prime order, alternating groups, groups of Lie type, and $26$ sporadic groups.''

\begin{tcolorbox}[
    enhanced,
    colback=blue!4,
    colframe=blue!55!black,
    colbacktitle=blue!18,
    coltitle=black,
    boxrule=0.8pt,
    arc=2mm,
    left=6pt,
    right=6pt,
    top=6pt,
    bottom=6pt,
    title=\textbf{{\LiveMathematicianBench} Example (Theorem \& Sketch)},
]
\footnotesize

\textbf{Original Theorem} (\cite{giesler2026jacobianringsinfinitesimaltorelli}) \par\smallskip
\label{Theorem_formula_ker_ITT_intro}
Let $\Delta$ be an $n$-dimensional lattice polytope with $\operatorname{Int}(\Delta) \cap M \nsubseteq \text{ hyperplane}$. Then
\begin{align*}
    \ker(d \mathcal{P}_{B,f}^k) \overset{\operatorname{mod} \operatorname{Lie}(T)}{=}&  \Big\langle g_{\Gamma}(f) \cdot x^w \in L(\Delta)/\mathbb{C} \cdot f \quad \mid \quad  \Gamma \leq \Delta \text{ a facet}, \\
    & w + v \in \big( \operatorname{Int}((k+1) \cdot \Delta) \cup \operatorname{Int}((k+1) \cdot \Gamma) \big) \cap M, \\ 
    & \forall \, v \in \operatorname{Int}(k \cdot \Delta) \cap M \Big\rangle.
\end{align*}
where $\langle \rangle$ denotes the span as $\mathbb{C}$-vector space.

\vspace{1em}
\hrule
\vspace{1em}

\textbf{Proof Sketch} \par\smallskip
This direct approach to the diagram of Griffiths
\[
\begin{tikzcd} \label{Commutative_diagram_Kodaira_Spencer_map_dP_intro}
    T_{B,f}  \arrow[swap]{dr}{d \mathcal{P}_{B,f}^k } \arrow{r}{\kappa_{f}} & H^{1}(Y_f,T_{Y_f})  \arrow{d}{\Phi_f^k} \\
    & \operatorname{Hom}(H^{k}(Y_f, \Omega_{Y_f}^{n-1-k}), H^{k+1}(Y_f, \Omega_{Y_f}^{n-2-k}))
\end{tikzcd}
\]
``yields a both computational and much explicit perspective.'' Under the assumptions of the main theorem, one ``easily deduce[s]
\[
\ker(d \mathcal{P}_{B,f}^k)=\ker(d \mathcal{P}_{B,f}^1),\quad k\ge 1
\]
We record the precise statement as follows.

\smallskip
\noindent
\textbf{Corollary} \label{corollary_conjecture_ker_dphi_f_k}\textbf{.}
\textit{Under the assumption of theorem we have}
\[
    \ker(d \phi_{f}) = \ker(d \phi_{f}^1).
\]

For an element $g=g_{\Gamma}(f)\cdot x^w\in\ker(d\mathcal{P}_f^1)$ one obtains the distinction
\[
\langle w,n_{\Gamma}\rangle=\begin{cases}
0 &\Leftrightarrow\ g\equiv 0,\\
-1 &\Leftrightarrow\ g\in\ker(\kappa_f),\\
\le -2 &\text{exceptional cases,}
\end{cases}
\]
and also ``$\langle w,n_{\Gamma'}\rangle\ge 0$ for $\Gamma'\neq\Gamma$ by the main theorem.'' The ``last case'' $\langle w,n_{\Gamma}\rangle\le -2$ ``causes $\Phi_{f\vert{\operatorname{Im}\,\kappa_f}}$ to be not injective, which implies in particular that the infinitesimal Torelli theorem (ITT) fails.''

For $n=3$, to rule out the exceptional case and conclude injectivity, the text sketches: ``we construct a $3$-dimensional empty polytope $Q$ with $6$ vertices,'' namely ``the convex span of an empty triangle with vertex $(0,0,0)$ and this triangle dilated by $w$,'' and then ``use a theorem of White on empty $3$-simplices to deduce $|Q\cap M|\ge 7$ under the assumption $\langle w,n_{\Gamma}\rangle\le -2$, contradiction.''

\end{tcolorbox}
\begin{tcolorbox}[
    enhanced,
    colback=green!4,
    colframe=green!55!black,
    colbacktitle=green!18,
    coltitle=black,
    boxrule=0.8pt,
    arc=2mm,
    left=6pt,
    right=6pt,
    top=6pt,
    bottom=6pt,
    title=\textbf{{\LiveMathematicianBench} Example  \refstepcounter{promptbox}\thepromptbox~ (Question \& Answers)\label{box:classification}},
]
\footnotesize

\textbf{Generated Question} \par\smallskip
Let $M\cong \mathbb Z^n$ be a lattice, $T=\operatorname{Spec}\mathbb C[M]\cong (\mathbb C^*)^n$, and let $f=\sum_{m\in \Delta\cap M} a_m x^m$ be a nondegenerate Laurent polynomial with $n$-dimensional Newton polytope $\Delta$. Write $L(\Delta):=\bigoplus_{m\in \Delta\cap M}\mathbb C\,x^m$, and let $\operatorname{Int}(\cdot)$ denote interior (relative interior on faces such as facets $\Gamma$). Assume that $\operatorname{Int}(\Delta)\cap M$ is not contained in any hyperplane. In the notation where $d\mathcal P_{B,f}^k$ is the differential under consideration and $g_{\Gamma}(f)$ is the element associated to a facet $\Gamma\le \Delta$, which subspace gives $\ker(d\mathcal P_{B,f}^k)$ modulo $\operatorname{Lie}(T)$?

\vspace{1em}
\hrule
\vspace{1em}

\textbf{Answer Choices} \par\smallskip
\textbf{A.} Modulo $\operatorname{Lie}(T)$, it is exactly the $\mathbb C$-span of all elements $g_{\Gamma}(f)\,x^w\in L(\Delta)/\mathbb C\cdot f$ such that $\Gamma\le \Delta$ is a facet and $w+v\in\big(\operatorname{Int}((k+1)\Delta)\cup \operatorname{Int}((k+1)\Gamma)\big)\cap M$ for every $v\in \operatorname{Int}(k\Delta)\cap M$. \textcolor{green!60!black}{\checkmark}

\textbf{B.} Modulo $\operatorname{Lie}(T)$, it is exactly the $\mathbb C$-span of all elements $g_{\Gamma}(f)\,x^w\in L(\Delta)/\mathbb C\cdot f$ such that $\Gamma\le \Delta$ is a facet and there exists $v\in \operatorname{Int}(k\Delta)\cap M$ with $w+v\in\big(\operatorname{Int}((k+1)\Delta)\cup \operatorname{Int}((k+1)\Gamma)\big)\cap M$.

\textbf{C.} Modulo $\operatorname{Lie}(T)$, it is contained in the $\mathbb C$-span of all elements $g_{\Gamma}(f)\,x^w\in L(\Delta)/\mathbb C\cdot f$ such that $\Gamma\le \Delta$ is a facet and $w+v\in\big(\operatorname{Int}((k+1)\Delta)\cup ((k+1)\Gamma)\big)\cap M$ for every $v\in \operatorname{Int}(k\Delta)\cap M$.

\textbf{D.} Modulo $\operatorname{Lie}(T)$, it is exactly the $\mathbb C$-span of all elements $g_{\Gamma}(f)\,x^w\in L(\Delta)/\mathbb C\cdot f$ such that $\Gamma\le \Delta$ is a facet and $w+v\in\big(\operatorname{Int}((k+1)\Delta)\cup \operatorname{Int}((k+1)\Gamma)\big)\cap M$ for every $v\in k\Delta\cap M$.

\textbf{E.} Modulo $\operatorname{Lie}(T)$, it is exactly the $\mathbb C$-span of all elements $g_{\Gamma}(f)\,x^w\in L(\Delta)/\mathbb C\cdot f$ such that $\Gamma\le \Delta$ is a facet and $w+v\in\big(\operatorname{Int}((k+1)\Delta)\cup \operatorname{Int}((k+1)\Gamma)\big)\cap M$ for every $v\in \operatorname{Int}(k\Gamma)\cap M$.

\end{tcolorbox}

\newpage
\paragraph{5. Existence.}
\emph{Canonical form:} $\exists\, x$ such that $P(x)$.
This category covers theorems guaranteeing that at least one object with a given property exists, without claiming uniqueness.
\emph{Example:} ``There exists a continuous, nowhere-differentiable function on $[0,1]$.''

\begin{tcolorbox}[
    enhanced,
    colback=blue!4,
    colframe=blue!55!black,
    colbacktitle=blue!18,
    coltitle=black,
    boxrule=0.8pt,
    arc=2mm,
    left=6pt,
    right=6pt,
    top=6pt,
    bottom=6pt,
    title=\textbf{{\LiveMathematicianBench} Example (Theorem \& Sketch)},
]
\footnotesize

\textbf{Original Theorem} (\cite{mejia2025mcduffsuperrigiditygroupii1}) \par\smallskip
\label{Thm:main}
There exists a family of countable $W^\ast$-McDuff groups $\{ G_i\}_{i\in I}$ such that $|I|=2^{\aleph_0}$, $G_i\not\cong G_j$ whenever $i\ne j$, and every $G_i$ is McDuff superrigid.

\vspace{1em}
\hrule
\vspace{1em}

\textbf{Proof Sketch} \par\smallskip
The groups $G_i$ are described as ``infinite direct sums of $W^\ast$-superrigid groups with property $(T)$ constructed in I. Chifan, A. Ioana, A. O. Sasyk, and D. Stef\u{a}nescu, CIOS21.'' Beyond this, the text only states that ``[t]he proof of Theorem To prove the main theorem, incorporates several new ideas on both the group-theoretic and analytic sides'' without outlining any steps or structure.

\end{tcolorbox}
\begin{tcolorbox}[
    enhanced,
    colback=green!4,
    colframe=green!55!black,
    colbacktitle=green!18,
    coltitle=black,
    boxrule=0.8pt,
    arc=2mm,
    left=6pt,
    right=6pt,
    top=6pt,
    bottom=6pt,
    title=\textbf{{\LiveMathematicianBench} Example  \refstepcounter{promptbox}\thepromptbox~ (Question \& Answers)\label{box:existence}},
]
\footnotesize

\textbf{Generated Question} \par\smallskip
A countable discrete group $G$ is called a $W^*$-McDuff group if its group von Neumann algebra $L(G)$ is a McDuff II$_1$ factor. Here, $G$ is called McDuff superrigid in the strongest sense if whenever $L(G)\cong L(H)$ for some group $H$, one has $H\cong G\times A$ for some ICC amenable group $A$ (ICC means every nontrivial conjugacy class is infinite). Which is the strongest existence statement that holds for countable $W^*$-McDuff groups?

\vspace{1em}
\hrule
\vspace{1em}

\textbf{Answer Choices} \par\smallskip
\textbf{A.} There exists a family $\{G_i\}_{i\in I}$ of countable $W^*$-McDuff groups with $|I|=2^{\aleph_0}$ such that $G_i\not\cong G_j$ whenever $i\ne j$, and every $G_i$ is McDuff superrigid. \textcolor{green!60!black}{\checkmark}

\textbf{B.} There exists a family $\{G_i\}_{i\in I}$ of countable $W^*$-McDuff groups with $|I|=2^{\aleph_0}$ such that $G_i\not\cong G_j$ whenever $i\ne j$, and every $G_i$ is McDuff superrigid in the strongest sense.

\textbf{C.} There exists a countably infinite family $\{G_i\}_{i\in I}$ of pairwise non-isomorphic countable $W^*$-McDuff groups such that every $G_i$ is McDuff superrigid.

\textbf{D.} There exists a family $\{G_i\}_{i\in I}$ of countable ICC groups with $|I|=2^{\aleph_0}$ such that $G_i\not\cong G_j$ whenever $i\ne j$, and every $G_i$ is McDuff superrigid.

\textbf{E.} For every countable $W^*$-McDuff group $G$, there exists a family $\{G_i\}_{i\in I}$ of countable $W^*$-McDuff groups with $|I|=2^{\aleph_0}$ such that $G_i\not\cong G_j$ whenever $i\ne j$, every $G_i$ is McDuff superrigid, and $G$ is isomorphic to one of the $G_i$.

\end{tcolorbox}

\newpage
\paragraph{6. Existential--Universal.}
\emph{Canonical form:} $\exists\, x\; \forall\, y,\; P(x,y)$.
This category covers theorems asserting the existence of an object (often a constant, function, or parameter) that ``works'' uniformly for all elements of a given class. The existential quantifier precedes the universal one.
\emph{Example:} ``There exists a constant $C > 0$ such that for all $f \in L^{2}(\mathbb{R})$, $\|f\|_{\infty} \leq C\,\|f\|_{H^{1}}$.''

\begin{tcolorbox}[
    enhanced,
    colback=blue!4,
    colframe=blue!55!black,
    colbacktitle=blue!18,
    coltitle=black,
    boxrule=0.8pt,
    arc=2mm,
    left=6pt,
    right=6pt,
    top=6pt,
    bottom=6pt,
    title=\textbf{{\LiveMathematicianBench} Example (Theorem \& Sketch)},
]
\footnotesize

\textbf{Original Theorem} (\cite{kala2026consecutivecubicfieldslarge}) \par\smallskip
\label{main:thm}
Fix a sign choice $\pm$, $\varepsilon>0$, and positive integers $k, H$. There exists a positive constant $c$ (depending on $\pm, \varepsilon, k, H$) such that for all sufficiently large $X$, there are at least $cX^{1-\kappa-\varepsilon}$ positive integers $d\leq X$ such that the following hold:
\begin{enumerate}
    \item For each $1\leq i\leq k$, if there exists a cubic field $K$ with $\pm\Delta_K=d+i$, then $h(K)>H$.

    \item We have 
    \[
    \#\{K \text{ cubic field }\mid \pm \Delta_K\in\{d+1,d+2,\dots,d+k\}\}\geq \frac{k}{2}.
    \]
\end{enumerate}

\vspace{1em}
\hrule
\vspace{1em}

\textbf{Proof Sketch} \par\smallskip
We will prove this result in \textbf{Theorem} \label{thm:final}\textbf{.}
\textit{Assume Setting (S). There exists a positive constant $c$ (depending on $\pm, \varepsilon, k, H$) such that for all sufficiently large $X$, there are at least $cX^{1-\kappa-\varepsilon}$ positive integers $t\leq X$ such that the following hold:
\begin{enumerate}
    \item[(1)] For each $1\leq i\leq k$, if there exists a cubic field $K$ with $\pm\Delta_K=a+tm+i$, then $h(K)>H$.
    \item[(2)] We have
    \[
    \#\{K \text{ cubic field }\mid \pm \Delta_K\in\{a+tm+1,a+tm+2,\dots,a+tm+k\}\}\geq \frac{k}{2}.
    \]
\end{enumerate}}
, and the proof is described as constructive. The approach is:
\begin{itemize}
    \item First, ``construct integers $0\leq a< m$ such that whenever there exists a cubic field $K$ with $\pm\Delta_K=d+i$, where $d\equiv a\pmod m$, then $h(K)>H$.'' This is obtained ``by controlling the \textit{genus number} which gives a lower bound for the size of the 2-torsion in the class group of $K$.'' The motivation given is that this ``sidesteps the difficult issue of estimating the regulator'' (so there is no need to ``restrict ourselves to a suitable family of cubic fields''), though consequently ``we only show that the class numbers are larger than some prescribed bound.''
    \item Then, ``show that indeed there are many cubic fields with $ \pm \Delta_K\in\{d+1,d+2,\dots,d+k\}$, which also ensures that our first assertion is not vacuous.''
\end{itemize}
It is also noted that, in counting the integers $d\le X$ in establishing the main theorem, one must ``switch between counting cubic fields and cubic discriminants,'' and that ``non-isomorphic cubic fields can have the same discriminant,'' i.e. discriminants are counted with multiplicity when counting fields.

\end{tcolorbox}
\begin{tcolorbox}[
    enhanced,
    colback=green!4,
    colframe=green!55!black,
    colbacktitle=green!18,
    coltitle=black,
    boxrule=0.8pt,
    arc=2mm,
    left=6pt,
    right=6pt,
    top=6pt,
    bottom=6pt,
    title=\textbf{{\LiveMathematicianBench} Example  \refstepcounter{promptbox}\thepromptbox~ (Question \& Answers)\label{box:exiuni}},
]
\footnotesize

\textbf{Generated Question} \par\smallskip
For a cubic field (that is, a degree-3 number field) $K$, let $\Delta_K$ denote its discriminant and $h(K)$ its class number. Fix once and for all either the plus sign or the minus sign, so that $\pm\Delta_K$ means $+\Delta_K$ throughout or $-\Delta_K$ throughout according to that choice; also fix a real number $\varepsilon>0$ and positive integers $k$ and $H$, and let $\kappa$ be the fixed constant appearing in the exponent below. Which is the strongest uniform existence statement that holds for all sufficiently large $X$?

\vspace{1em}
\hrule
\vspace{1em}

\textbf{Answer Choices} \par\smallskip
\textbf{A.} One of the remaining options is correct, but a stronger result can be proven. \textcolor{green!60!black}{\checkmark}

\textbf{B.} There exists a positive constant $c$ depending on $\pm,\varepsilon,k,H$ such that for all sufficiently large $X$, there are at least $cX^{1-\kappa-\varepsilon}$ positive integers $d\le X$ such that the following hold: for each $1\le i\le k$, if there exists a cubic field $K$ with $\pm\Delta_K=d+i$, then $h(K)>H$; and moreover
\[
\#\{n\in\{d+1,d+2,\dots,d+k\}\mid n\text{ is a cubic discriminant with sign }\pm\}\ge \frac{k}{2}.
\]

\textbf{C.} There exists a positive constant $c$ depending on $\pm,\varepsilon,k,H$ such that for all sufficiently large $X$, there are at least $cX^{1-\kappa-\varepsilon}$ positive integers $d\le X$ for which, for each $1\le i\le k$, if there exists a cubic field $K$ with $\pm\Delta_K=d+i$, then $h(K)>H$.

\textbf{D.} There exists a positive constant $c$ depending on $\pm,\varepsilon,k,H$ such that for all sufficiently large $X$, there are at least $cX^{1-\kappa-\varepsilon}$ positive integers $d\le X$ such that the following hold: for each $1\le i\le k$, if there exists a cubic field $K$ with $\pm\Delta_K=d+i$, then $h(K)>H$; and moreover
\[
\#\{K\text{ cubic field}\mid \pm\Delta_K\in\{d+1,d+2,\dots,d+k\}\}\ge \frac{k+1}{2}.
\]

\textbf{E.} There exists a positive constant $c$ depending on $\pm,\varepsilon,k,H$ such that for all sufficiently large $X$, there are at least $cX^{1-\kappa}$ positive integers $d\le X$ such that the following hold: for each $1\le i\le k$, if there exists a cubic field $K$ with $\pm\Delta_K=d+i$, then $h(K)>H$; and moreover
\[
\#\{K\text{ cubic field}\mid \pm\Delta_K\in\{d+1,d+2,\dots,d+k\}\}\ge \frac{k}{2}.
\]

\end{tcolorbox}

\newpage
\paragraph{7. Implication.}
\emph{Canonical form:} $A \implies B$, or ``if $P$ holds, then $Q$ holds.''
This category covers theorems in which a set of hypotheses implies a specific conclusion, without asserting the converse.
\emph{Example:} ``If $f$ is a continuous function on a closed interval $[a,b]$, then $f$ is uniformly continuous on $[a,b]$.''

\begin{tcolorbox}[
    enhanced,
    colback=blue!4,
    colframe=blue!55!black,
    colbacktitle=blue!18,
    coltitle=black,
    boxrule=0.8pt,
    arc=2mm,
    left=6pt,
    right=6pt,
    top=6pt,
    bottom=6pt,
    title=\textbf{{\LiveMathematicianBench} Example (Theorem \& Sketch)},
]
\footnotesize

\textbf{Original Theorem} (\cite{moraga2026degenerationsclustertypevarieties}) \par\smallskip
\label{thm:toric-deg}
Let $\pi\colon \mathcal{X}\rightarrow \mathbb{D}$ be a projective fibration.
Let $(\mathcal{X},\mathcal{X}_0+\mathcal{B})$ be a log Calabi--Yau pair over $\mathbb{D}$ for which $(\mathcal{X},\mathcal{X}_0)$ is plt. 
If $(\mathcal{X}_t,\mathcal{B}_t)$ is a toric pair\footnote{Meaning that $\mathcal{X}_t$ is a projective toric variety and $\mathcal{B}_t$ is the reduced torus invariant boundary.} for $t \in \mathbb{D}^*$,
then $(\mathcal{X}_0,\mathcal{B}_0)$ is a finite quotient of a toric pair.

\vspace{1em}
\hrule
\vspace{1em}

\textbf{Proof Sketch} \par\smallskip
In the proof of the main theorem, we will see that $(\mathcal{X},\mathcal{B})\rightarrow \mathbb{D}$ is indeed a finite quotient of an isotrivial toric family. Thus, in the toric case, there are not many interesting degenerations for which the degeneration of the toric boundary still has lc singularities.

\end{tcolorbox}
\begin{tcolorbox}[
    enhanced,
    colback=green!4,
    colframe=green!55!black,
    colbacktitle=green!18,
    coltitle=black,
    boxrule=0.8pt,
    arc=2mm,
    left=6pt,
    right=6pt,
    top=6pt,
    bottom=6pt,
    title=\textbf{\LiveMathematicianBench Example  \refstepcounter{promptbox}\thepromptbox~ (Question \& Answers)\label{box:implication}},
]
\footnotesize

\textbf{Generated Question} \par\smallskip
Let $\pi\colon \mathcal{X}\to \mathbb{D}$ be a projective morphism to the unit disk $\mathbb{D}$, with central fiber $\mathcal{X}_0=\pi^{-1}(0)$. Let $\mathcal{B}$ be a divisor such that $(\mathcal{X},\mathcal{X}_0+\mathcal{B})$ is a log Calabi--Yau pair over $\mathbb{D}$ (so $K_{\mathcal{X}}+\mathcal{X}_0+\mathcal{B}$ is trivial over $\mathbb{D}$), and assume that $(\mathcal{X},\mathcal{X}_0)$ is plt (purely log terminal). For each $t\in \mathbb{D}^*=\mathbb{D}\setminus\{0\}$, write $\mathcal{X}_t=\pi^{-1}(t)$ and $\mathcal{B}_t=\mathcal{B}|_{\mathcal{X}_t}$. Suppose that for every $t\in \mathbb{D}^*$, the fiber pair $(\mathcal{X}_t,\mathcal{B}_t)$ is toric, meaning that $\mathcal{X}_t$ is a projective toric variety and $\mathcal{B}_t$ is its reduced torus-invariant boundary. Under these assumptions, which is the strongest statement about the special fiber pair $(\mathcal{X}_0,\mathcal{B}_0)$ that holds?

\vspace{1em}
\hrule
\vspace{1em}

\textbf{Answer Choices} \par\smallskip
\textbf{A.} The special fiber pair $(\mathcal{X}_0,\mathcal{B}_0)$ is a finite quotient of a toric pair. \textcolor{green!60!black}{\checkmark}

\textbf{B.} The special fiber pair $(\mathcal{X}_0,\mathcal{B}_0)$ becomes a toric pair after normalizing $\mathcal{X}_0$ and replacing $\mathcal{B}_0$ by the induced reduced boundary on the normalization.

\textbf{C.} The special fiber pair $(\mathcal{X}_0,\mathcal{B}_0)$ is birational to a toric pair.

\textbf{D.} After a finite base change of the disk, the pulled-back degeneration has toric central fiber; equivalently, $(\mathcal{X}_0,\mathcal{B}_0)$ becomes toric after such a base change.

\textbf{E.} There exists a finite morphism of pairs $(\mathcal{X}_0,\mathcal{B}_0)\to (Y,D)$ onto a toric pair $(Y,D)$.

\end{tcolorbox}

\newpage
\paragraph{8. Independence / Consistency.}
\emph{Canonical form:} $T \not\vdash P$ and $T \not\vdash \neg P$, or ``statement $P$ is independent of axiom system $T$.''
This category covers theorems expressing a meta-mathematical statement that a proposition cannot be proved or refuted from given axioms.
\emph{Example:} ``The Continuum Hypothesis is independent of ZFC.''

\begin{tcolorbox}[
    enhanced,
    colback=blue!4,
    colframe=blue!55!black,
    colbacktitle=blue!18,
    coltitle=black,
    boxrule=0.8pt,
    arc=2mm,
    left=6pt,
    right=6pt,
    top=6pt,
    bottom=6pt,
    title=\textbf{{\LiveMathematicianBench} Example (Theorem \& Sketch)},
]
\footnotesize

\textbf{Original Theorem} (\cite{kaplan2025noteketonenorderlipschitz}) \par\smallskip
\label{theorem: intro}
Consistently from a measurable cardinal, there are two $\sigma$-complete ultrafilters $\mathcal{V}, \mathcal{W}$ such that $\mathcal{V}$ is Lipschitz below $\mathcal{W}$, but $\mathcal{V}$ and $\mathcal{W}$ are Ketonen-incomparable. Furthermore, the same is also consistent with the assumption that the Weak Ultrapower Axiom holds.

\vspace{1em}
\hrule
\vspace{1em}

\textbf{Proof Sketch} \par\smallskip
We will rely on the construction from \cite{kaplan2025number} in the proof of the main theorem. The paper then proceeds by: (1) defining the Ketonen and Lipschitz orders and outlining their basic properties; (2) showing, consistently, that the Ketonen and Lipschitz orders do not coincide (not in a model of the Weak UA). We first prove the following theorem. 

\smallskip
\noindent
\textbf{Theorem} \label{Theorem: Separating Lipschitz from Ketonen the Easy way}\textbf{.}
\textit{Assume $V= L[U]$, $\kappa$ is the unique measurable cardinal, $\po$ is the forcing notion described above and $G,G^*, U_0,U_1$ are as above. Denote $\vc = U_0$ and $\wc$ the measure derived from $j^{V[G]}_{(U_1)^2}$ using the ordinal $j_U(\kappa)+\kappa$ as a seed. Then $\vc<_{L} \wc$ but $\vc, \wc$ are Ketonen incomparable.}

(3) sketching basic properties of forcing with nonstationary support products; and (4) separating the Ketonen and Lipschitz orders in a model of the Weak UA, thereby completing the proof of the main theorem.

\end{tcolorbox}
\begin{tcolorbox}[
    enhanced,
    colback=green!4,
    colframe=green!55!black,
    colbacktitle=green!18,
    coltitle=black,
    boxrule=0.8pt,
    arc=2mm,
    left=6pt,
    right=6pt,
    top=6pt,
    bottom=6pt,
    title=\textbf{{\LiveMathematicianBench} Example  \refstepcounter{promptbox}\thepromptbox~ (Question \& Answers)\label{box:independence}},
]
\footnotesize

\textbf{Generated Question} \par\smallskip
Assume the consistency strength of a measurable cardinal. A $\sigma$-complete ultrafilter is an ultrafilter closed under countable intersections. Using the standard comparison notions $<_L$ (the Lipschitz order) and $<_k$ (the Ketonen order) on $\sigma$-complete ultrafilters, where ``Ketonen-incomparable'' means that neither ultrafilter is below the other in $<_k$, and writing WUA for the Weak Ultrapower Axiom, which is the strongest consistency statement that holds about $\sigma$-complete ultrafilters?

\vspace{1em}
\hrule
\vspace{1em}

\textbf{Answer Choices} \par\smallskip
\textbf{A.} Relative to a measurable cardinal, it is consistent that there exist two $\sigma$-complete ultrafilters $\mathcal V$ and $\mathcal W$ such that $\mathcal V$ is Lipschitz below $\mathcal W$, while $\mathcal V$ and $\mathcal W$ are Ketonen-incomparable; moreover, it is also consistent that such a situation occurs together with the Weak Ultrapower Axiom. \textcolor{green!60!black}{\checkmark}

\textbf{B.} Relative to a measurable cardinal, it is consistent that there exist two $\sigma$-complete ultrafilters $\mathcal V$ and $\mathcal W$ such that $\mathcal V$ is Lipschitz below $\mathcal W$, while $\mathcal V$ and $\mathcal W$ are Ketonen-incomparable; however, no model satisfying the Weak Ultrapower Axiom can contain such a pair.

\textbf{C.} Relative to a measurable cardinal, it is consistent that there exist two $\sigma$-complete ultrafilters $\mathcal V$ and $\mathcal W$ such that $\mathcal V$ is Lipschitz below $\mathcal W$, while $\mathcal V$ and $\mathcal W$ are Ketonen-incomparable.

\textbf{D.} Relative to a measurable cardinal, it is consistent that there exist two $\sigma$-complete ultrafilters $\mathcal V$ and $\mathcal W$ such that $\mathcal V$ is Lipschitz below $\mathcal W$ and the Lipschitz and Ketonen orders do not coincide on $\sigma$-complete ultrafilters; nevertheless, in every model of the Weak Ultrapower Axiom, any relation $\mathcal V<_L\mathcal W$ already implies that $\mathcal V$ and $\mathcal W$ are comparable in the Ketonen order.

\textbf{E.} Relative to a measurable cardinal, it is consistent that the Lipschitz and Ketonen orders on $\sigma$-complete ultrafilters can fail to coincide, but only outside the Weak Ultrapower Axiom; under WUA, the Ketonen order is always linear on $\sigma$-complete ultrafilters.

\end{tcolorbox}

\newpage
\paragraph{9. Inequality / Bound.}
\emph{Canonical form:} $f(x) \leq g(x)$ (or $\geq$, $<$, $>$), holding universally.
This category covers theorems providing a quantitative estimate, such as a uniform bound, norm inequality, or comparison between quantities.
\emph{Example:} ``For all $f \in L^{1}(\mathbb{R}) \cap L^{2}(\mathbb{R})$, $\|\hat{f}\|_{2} = \|f\|_{2}$.''

\begin{tcolorbox}[
    enhanced,
    colback=blue!4,
    colframe=blue!55!black,
    colbacktitle=blue!18,
    coltitle=black,
    boxrule=0.8pt,
    arc=2mm,
    left=6pt,
    right=6pt,
    top=6pt,
    bottom=6pt,
    title=\textbf{{\LiveMathematicianBench} Example (Theorem \& Sketch)},
]
\footnotesize

\textbf{Original Theorem} (\cite{ozawa2026crossingnumbersknotsclosed}) \par\smallskip
\textbf{Theorem} (Fundamental inequality)\label{thm:main_intro}\textbf{.}
Let $F\subset S^3$ be a closed surface. Then every knot $K\subset S^3$ satisfies:
\[
c(K;F) \ge 2\bigl(t(K)-\delta(F)\bigr)+1,
\]
where $t(K)$ is the tunnel number of $K$, and $\delta(F)=g(M_1)+g(M_2)-g(F)$ is the Heegaard deficiency of $F$.

\vspace{1em}
\hrule
\vspace{1em}

\textbf{Proof Sketch} \par\smallskip
To prove the main theorem, the argument ``relies on a chain of inequalities relating the surface crossing number $c(K;F)$, the surface ascending number $a(K;F)$, and the surface bridge number $b(K;F)$'':
\[
\frac{c(K;F)-1}{2} \ge a(K;F) \ge b(K;F)-1 \ge t(K)-\delta(F).
\]
This sequence yields the stated lower bound for $c(K;F)$ in terms of ``a combination of the knot exterior's complexity, $t(K)$, and the ambient surface's Heegaard deficiency, $\delta(F)$.''

For optimality, the introduction says they ``prove that the linear lower bound in the main theorem is asymptotically optimal'' by taking iterated connected sums $K_m$ of a prime knot and obtaining
\[
c(K_m;F) \ge 2m - 2\delta(F) + 1.
\]
``Pairing this with a general upper bound derived from the subadditivity of the planar crossing number yields $c(K_m;F)=\Theta(t(K_m))$,'' so ``no general lower bound of higher order exists,'' and $c(K;F)$ grows linearly with tunnel number.

\end{tcolorbox}
\begin{tcolorbox}[
    enhanced,
    colback=green!4,
    colframe=green!55!black,
    colbacktitle=green!18,
    coltitle=black,
    boxrule=0.8pt,
    arc=2mm,
    left=6pt,
    right=6pt,
    top=6pt,
    bottom=6pt,
    title=\textbf{{\LiveMathematicianBench} Example  \refstepcounter{promptbox}\thepromptbox~ (Question \& Answers)\label{box:inequality}},
]
\footnotesize

\textbf{Generated Question} \par\smallskip
Let $F\subset S^3$ be a closed surface, and let $M_1$ and $M_2$ be the two components of $S^3\setminus F$. Define the Heegaard deficiency of $F$ by
\[
\delta(F)=g(M_1)+g(M_2)-g(F),
\]
where $g(M_i)$ denotes the Heegaard genus of $M_i$ and $g(F)$ denotes the genus of $F$. For a knot $K\subset S^3$, let $c(K;F)$ denote the surface crossing number of $K$ with respect to $F$, i.e. the minimal number of crossings among all regular diagrams of $K$ obtained by isotoping $K$ into a regular neighborhood of $F$. If $t(K)$ is the tunnel number of $K$, which is the strongest quantitative estimate that holds for $c(K;F)$?

\vspace{1em}
\hrule
\vspace{1em}

\textbf{Answer Choices} \par\smallskip
\textbf{A.} For every knot $K\subset S^3$, 
\[
c(K;F)\ge 2\bigl(t(K)-\delta(F)\bigr)+1. \textcolor{green!60!black}{\checkmark}
\]

\textbf{B.} For every knot $K\subset S^3$,
\[
c(K;F)\ge 2\bigl(t(K)-\delta(F)+1\bigr)+1.
\]

\textbf{C.} For every knot $K\subset S^3$,
\[
c(K;F)\ge 2\bigl(t(K)-\delta(F)\bigr).
\]

\textbf{D.} For every knot $K\subset S^3$,
\[
c(K;F)\ge 2t(K)-\delta(F)+1.
\]

\textbf{E.} For every knot $K\subset S^3$,
\[
c(K;F)\ge 2\bigl(t(K)-\delta(F)\bigr)+2.
\]

\end{tcolorbox}

\newpage
\paragraph{10. Nonexistence.}
\emph{Canonical form:} $\not\exists\, x$ such that $P(x)$ (or equivalently, $\forall\, x,\; \neg P(x)$).
This category covers theorems asserting that no object satisfies a given property.
\emph{Example:} ``There does not exist a rational number whose square is $2$.''

\begin{tcolorbox}[
    enhanced,
    colback=blue!4,
    colframe=blue!55!black,
    colbacktitle=blue!18,
    coltitle=black,
    boxrule=0.8pt,
    arc=2mm,
    left=6pt,
    right=6pt,
    top=6pt,
    bottom=6pt,
    title=\textbf{{\LiveMathematicianBench} Example (Theorem \& Sketch)},
]
\footnotesize

\textbf{Original Theorem} (\cite{wang2025nonnegativescalarcurvaturespin}) \par\smallskip
\label{thm:connectedSum}
Let $M^n$ be a closed aspherical manifold and $N$ a spin manifold. Assume that the rational strong Novikov conjecture holds for $\pi_1(M)$.
Then the connected sum $M\# N$ does not carry any complete metric with positive scalar curvature. Furthermore, any complete metric with non-negative scalar curvature on $M\# N$ is Ricci-flat.

\vspace{1em}
\hrule
\vspace{1em}

\textbf{Proof Sketch} \par\smallskip
Our proofs rely on Dirac operator methods. A fundamental ingredient is the relative index theorem for higher indices due to Xie and Yu (Theorem~A in Xie--Yu, \emph{[title not provided in ref\_dict]} (2014)). Inspired by the approaches in Su--Zhang, \emph{[title not provided in ref\_dict]}, and Wang--Xie--Linf, \emph{[title not provided in ref\_dict]}, we construct an explicit representative of the relative higher index and compute it in the $K$-theory of the group $C^*$-algebra. We also provide a Dirac-operator-based proof of the Ricci-flatness statement, following ideas similar to those in Wang--Zhu, \emph{[title not provided in ref\_dict]}. (In establishing the main theorem, we treat it as a special case of

\smallskip
\noindent
\textbf{Theorem} \label{thm:main}\textbf{.}
\textit{Let $M^n$ be a closed aspherical manifold and $\Omega\subset M$ a smooth region. Assume that
\[\pi_1(\Omega)\to \pi_1(M)\textup{ is the trivial map}.\]
Let $N^n$ be a spin manifold with boundary, and $i\colon \partial N\to \partial \Omega$ a diffeomorphism preserving the spin structure\footnote{Although $M$ itself need not be spin, the triviality of $\pi_1(\Omega)\to\pi_1(M)$ implies that $\Omega$ and $\partial\Omega$ inherit a canonical spin structure from the spin universal cover $\widetilde M$.}. Set $Z := (M\backslash\Omega)\cup_{\partial\Omega} N$. If the rational strong Novikov conjecture holds for $\pi_1(M)$, then $Z$ does not carry any complete metric with positive scalar curvature. Furthermore, any complete metric with non-negative scalar curvature on $Z$ is Ricci-flat.}

by choosing $\Omega$ to be a small ball in $M$.)

\end{tcolorbox}
\begin{tcolorbox}[
    enhanced,
    colback=green!4,
    colframe=green!55!black,
    colbacktitle=green!18,
    coltitle=black,
    boxrule=0.8pt,
    arc=2mm,
    left=6pt,
    right=6pt,
    top=6pt,
    bottom=6pt,
    title=\textbf{{\LiveMathematicianBench} Example  \refstepcounter{promptbox}\thepromptbox~ (Question \& Answers)\label{box:nonexi}},
]
\footnotesize

\textbf{Generated Question} \par\smallskip
Let $M^n$ be a closed aspherical manifold (that is, a closed manifold whose universal cover is contractible), and let $N$ be a spin $n$-manifold so that the connected sum $M\# N$ is defined. Write $\Gamma=\pi_1(M)$, and assume that the rational strong Novikov conjecture holds for $\Gamma$, i.e. the assembly map $\mu\colon K_*(M)\otimes \mathbb{Q}\to K_*(C_r^*(\Gamma))\otimes \mathbb{Q}$ is injective. Which is the strongest statement about complete Riemannian metrics on $M\# N$ that is valid under these hypotheses?

\vspace{1em}
\hrule
\vspace{1em}

\textbf{Answer Choices} \par\smallskip
\textbf{A.} The connected sum $M\# N$ admits no complete metric of positive scalar curvature. Moreover, if $M\# N$ carries a complete metric of non-negative scalar curvature, then that metric must be Ricci-flat. \textcolor{green!60!black}{\checkmark}

\textbf{B.} The connected sum $M\# N$ admits no complete metric of positive scalar curvature. Moreover, every complete metric on $M\# N$ with non-negative scalar curvature is in fact flat.

\textbf{C.} Every complete metric on $M\# N$ with non-negative scalar curvature has identically zero scalar curvature; in particular, $M\# N$ admits no complete metric of positive scalar curvature.

\textbf{D.} The connected sum $M\# N$ admits no complete metric of non-negative scalar curvature at all; equivalently, $M\# N$ carries neither a complete Ricci-flat metric nor a complete metric of positive scalar curvature.

\textbf{E.} If $M\# N$ carries one complete metric of non-negative scalar curvature, then every complete metric on $M\# N$ must also have non-negative scalar curvature and be Ricci-flat.

\end{tcolorbox}

\newpage
\paragraph{11. Uniqueness.}
\emph{Canonical form:} $\exists!\, x$ such that $P(x)$.
This category covers theorems asserting both existence and uniqueness of an object satisfying a given property.
\emph{Example:} ``For every positive definite matrix $A$, there exists a unique positive definite matrix $B$ such that $B^{2} = A$.''

\begin{tcolorbox}[
    enhanced,
    breakable,
    colback=blue!4,
    colframe=blue!55!black,
    colbacktitle=blue!18,
    coltitle=black,
    boxrule=0.8pt,
    arc=2mm,
    left=6pt,
    right=6pt,
    top=6pt,
    bottom=6pt,
    title=\textbf{{\LiveMathematicianBench} Example (Theorem \& Sketch)},
]
\footnotesize

\textbf{Original Theorem} (\cite{mu2025globalexistencerelativisticvlasovpoisson}) \par\smallskip
\label{T1}
Let $f_{0} \in C^{1,\mu}_{0}\big(\bar{\Omega}\times\mathbb{R}^{2}\big), f_{0}\geq 0$ for some $0 <\mu< 1$, satisfying
\begin{align}
  f_{0}\in C^{1,\mu},\quad f_{0}(x,v)=\text{constant,\,\,  dist}((x,v), \Gamma)\leq \delta_{0},\label{v2.8}
\end{align}
Suppose that $h \in C^{2,\mu} (\partial\Omega)$ satisfies
\begin{align}
\int_{\Omega}f_{0}(x,v){\rm d}x{\rm d}v=\int_{\partial\Omega}h(x){\rm d}l, \label{v1.12}
\end{align}
and $h > 0$. Then there exists
a unique solution $f \in C^{1;1,\lambda}_{t;(x,v)} \big((0,\infty)\times \bar{\Omega}\times \mathbb{R}^{2}\big),
\varphi\in  C^{1;3,\lambda}_{t;x}\big((0,\infty)\times \bar{\Omega}\big) $
for
some $0 < \lambda < \mu$, of the relativistic Vlasov-Poisson system
\begin{align}
&\partial_t f+\hat{v}\cdot\nabla_{x}f+\nabla_{x}\varphi\cdot\nabla_{v}f=0, \label{v1.5}\\
&\Delta\varphi=\rho, \label{v1.6}
\end{align}
with compact support
in $x$ and $v$, where the  initial boundary conditions of $(f,\varphi)$ satisfy
\begin{align}
&f(0,x,v)=f_{0}(x,v),\quad x\in \Omega,\; v\in \mathbb{R}^{2}, \label{v1.8}\\
&f(t,x,v)=f(t,x,v^{*}),\quad x\in \partial\Omega, \; v\in \mathbb{R}^{2},\; t>0, \label{v1.9}
\end{align}
\begin{align}
 &f_{0}(x,v)\geq 0,  \label{v1.10}\\
 &v^{*}=v-2\big(n_{x}\cdot v\big)n_{x}, \quad (x,v)\in \partial\Omega \times \mathbb{R}^{2}, \label{v1.11}
 \end{align}
and
\begin{align}
   &\frac{\partial \varphi}{\partial n_{x}}=h(x),\,\,\, x\in \partial \Omega,\,t>0,\label{v1.7}
\end{align}
respectively.

\vspace{1em}
\hrule
\vspace{1em}

\textbf{Proof Sketch} \par\smallskip
To prove the main results (in particular the main theorem), the introduction describes the following strategy.

1. \textbf{Linearized well-posedness via a velocity lemma:} ``we first apply the velocity lemma to establish the well-posedness of linearized problems.''

2. \textbf{Iterative construction and convergence:} ``Then, we construct an iterative scheme and show the convergence of the iterative sequences.'' The ``main issues'' in this step are ``the uniform boundedness in a given function space and the prolongation of uniform estimates for the functions $f^{n}$.''

3. \textbf{Bootstrapping to final regularity/conclusion:} ``Finally, we use bootstrapping techniques to reach the desired conclusions.''

4. \textbf{Control of boundary interactions / collisions (via adapted coordinates):} Near the singular set and boundary, the plan is to ``draw on insights from previous research,'' assuming ``the initial data $f_{0}$ remains constant near the singular set,'' and ``to analyze the characteristic curves close to the boundary, we will use geometric methods.'' For the (Neumann) boundary analysis, they ``select coordinate variables $(\alpha,\beta)$,'' where ``$\alpha$ characterizes the distance from points on the characteristic curve to the singular set'' and ``$\beta$ \dots describes the number of collisions between particles and the boundary,'' with the observation that ``the number of collisions is inversely proportional to the distance to the singular set.'' For this purpose (to ensure the collision number is uniformly bounded), they assume
\[
  f_{0}\in C^{1,\mu},\quad f_{0}(x,v)=\text{constant,\,\,  dist}((x,v), \Gamma)\leq \delta_{0},
\]

5. \textbf{Paper roadmap matching the proof:} Sections 2--5 implement this plan: ``introduce an iterative system \dots define a sequence of functions $\{ f^{n}\}$, the limit \dots is the desired global solution,'' then ``establish the well-posedness of the linear problem,'' then ``show \dots convergence \dots under the condition that $Q(t)$ is bounded,'' and finally ``prove the boundedness of the function $Q(t)$. This concludes the proof of the first theorem.''

\end{tcolorbox}
\begin{tcolorbox}[
    enhanced,
    colback=green!4,
    colframe=green!55!black,
    colbacktitle=green!18,
    coltitle=black,
    boxrule=0.8pt,
    arc=2mm,
    left=6pt,
    right=6pt,
    top=6pt,
    bottom=6pt,
    title=\textbf{{\LiveMathematicianBench} Example  \refstepcounter{promptbox}\thepromptbox~ (Question \& Answers)\label{box:uniqueness}},
]
\footnotesize

\textbf{Generated Question} \par\smallskip
Let $0<\mu<1$, let $\Omega\subset \mathbb{R}^2$ have smooth boundary $\partial\Omega$, and let $n_x$ denote the outward unit normal at $x\in\partial\Omega$. Write $\Gamma:=\{(x,v)\in \partial\Omega\times\mathbb{R}^2: v\in T_x\partial\Omega\}$, let $C^{1,\mu}_0(\bar\Omega\times\mathbb{R}^2)$ denote $C^{1,\mu}$ functions with compact support, and let ${\rm d}l$ denote arc-length measure on $\partial\Omega$. Suppose $f_0\in C^{1,\mu}_0(\bar\Omega\times\mathbb{R}^2)$ is nonnegative and satisfies $f_0(x,v)=\text{constant}$ whenever $\operatorname{dist}((x,v),\Gamma)\le \delta_0$, and let $h\in C^{2,\mu}(\partial\Omega)$ satisfy $h>0$ and $\int_{\Omega} f_0(x,v)\,{\rm d}x\,{\rm d}v=\int_{\partial\Omega} h(x)\,{\rm d}l$. For the relativistic Vlasov-Poisson system
\[
\partial_t f+\hat v\cdot\nabla_x f+\nabla_x\varphi\cdot\nabla_v f=0,\qquad \Delta\varphi=\rho,\qquad \hat v=\frac{v}{\sqrt{1+|v|^2}},\quad \rho(t,x)=\int_{\mathbb{R}^2} f(t,x,v)\,{\rm d}v,
\]
with initial condition $f(0,x,v)=f_0(x,v)$, specular reflection boundary condition $f(t,x,v)=f(t,x,v^*)$ on $\partial\Omega\times\mathbb{R}^2\times(0,\infty)$, where $v^*=v-2(n_x\cdot v)n_x$, and Neumann boundary condition $\partial\varphi/\partial n_x=h(x)$ on $\partial\Omega\times(0,\infty)$, what is the strongest statement about existence and uniqueness that follows?

\vspace{1em}
\hrule
\vspace{1em}

\textbf{Answer Choices} \par\smallskip
\textbf{A.} There exists a unique pair $(f,\varphi)$ with $f\in C^{1;1,\lambda}_{t;(x,v)}((0,\infty)\times \bar\Omega\times \mathbb{R}^2)$ and $\varphi\in C^{1;3,\lambda}_{t;x}((0,\infty)\times \bar\Omega)$ for some $0<\lambda<\mu$, solving the above relativistic Vlasov-Poisson system, satisfying the stated initial, specular-reflection, and Neumann boundary conditions, and having compact support in $x$ and $v$. \textcolor{green!60!black}{\checkmark}

\textbf{B.} There exists a unique pair $(f,\varphi)$ with $f\in C^{1;1,\lambda}_{t;(x,v)}((0,\infty)\times \bar\Omega\times \mathbb{R}^2)$ and $\varphi\in C^{1;3,\lambda}_{t;x}((0,\infty)\times \bar\Omega)$ for some $0<\lambda<\mu$, solving the above relativistic Vlasov-Poisson system, satisfying the stated initial, specular-reflection, and Neumann boundary conditions, and having compact support in $x$ and $v$; moreover one may take $\lambda=\mu$.

\textbf{C.} There exists at least one pair $(f,\varphi)$ with $f\in C^{1;1,\lambda}_{t;(x,v)}((0,\infty)\times \bar\Omega\times \mathbb{R}^2)$ and $\varphi\in C^{1;3,\lambda}_{t;x}((0,\infty)\times \bar\Omega)$ for some $0<\lambda<\mu$, solving the above relativistic Vlasov-Poisson system, satisfying the stated initial, specular-reflection, and Neumann boundary conditions, and having compact support in $x$ and $v$.

\textbf{D.} There exists a unique pair $(f,\varphi)$ with $f\in C^{1;1,\lambda}_{t;(x,v)}((0,\infty)\times \bar\Omega\times \mathbb{R}^2)$ and $\varphi\in C^{1;3,\lambda}_{t;x}((0,\infty)\times \bar\Omega)$ for some $0<\lambda<\mu$, solving the above relativistic Vlasov-Poisson system, satisfying the stated initial, specular-reflection, and Neumann boundary conditions, and having compact support in $x$ and $v$, even without the flatness assumption $f_0(x,v)=\mathrm{constant}$ whenever $\operatorname{dist}((x,v),\Gamma)\le \delta_0$.

\textbf{E.} For every $0<\lambda<\mu$, there exists a unique pair $(f,\varphi)$ with $f\in C^{1;1,\lambda}_{t;(x,v)}((0,\infty)\times \bar\Omega\times \mathbb{R}^2)$ and $\varphi\in C^{1;3,\lambda}_{t;x}((0,\infty)\times \bar\Omega)$, solving the above relativistic Vlasov-Poisson system, satisfying the stated initial, specular-reflection, and Neumann boundary conditions, and having compact support in $x$ and $v$.

\end{tcolorbox}

\newpage
\paragraph{12. Universal.}
\emph{Canonical form:} $\forall\, x,\; P(x)$.
This category covers theorems asserting that a property holds for \emph{every} object in a specified class.
\emph{Example:} ``Every finite-dimensional real vector space admits an inner product.''

\begin{tcolorbox}[
    enhanced,
    colback=blue!4,
    colframe=blue!55!black,
    colbacktitle=blue!18,
    coltitle=black,
    boxrule=0.8pt,
    arc=2mm,
    left=6pt,
    right=6pt,
    top=6pt,
    bottom=6pt,
    title=\textbf{{\LiveMathematicianBench} Example (Theorem \& Sketch)},
]
\footnotesize

\textbf{Original Theorem} (\cite{grabbel2026symmetrieshyperboliclatticeslarge}) \par\smallskip
\label{thm:maintheorem}
Let $L$ be a hyperbolic lattice of rank at least $46$. Then, the exceptional lattice of $L$ is trivial.

\vspace{1em}
\hrule
\vspace{1em}

\textbf{Proof Sketch} \par\smallskip
To prove the main theorem, we start from Nikulin's explicit description of the exceptional lattice (Nikulin, ``Elliptic fibrations on K3 surfaces'', Theorem~4.2), see also the following proposition.

\smallskip
\noindent
\textbf{Proposition} \label{prop:exceptionalIntersection}\textbf{.}
\textit{Let $L$ be a lattice such that $\mathcal E_\infty(L)$ is non-empty. 
Then the exceptional lattice satisfies
\[
    E(L)=\bigcap _{e\in \mathcal E_{\infty}(L)} (e^{\perp})^{(2)}_{\mathrm{pr}}.
\]}

It implies that $E(L)$ is trivial ``as soon as every simple $(-2)$-root $r\in L$ admits a cusp with infinite stabilizer $e\in \overline{\mathcal{D}}_L$ such that $e.r>0$.''

For a hyperbolic lattice $L$ of rank at least $46$, one ``tak[es] an arbitrary maximal overlattice of $L$'' to reduce to lattices of the form
\[
L = U\oplus E_8^n\oplus M,\quad n\ge 5,\ \rk(M)\le 11
\]
by the following proposition.

\smallskip
\noindent
\textbf{Proposition} \label{prop:structure}\textbf{.}
\textit{If $L$ is a hyperbolic lattice of rank at least $6$, then there exist an overlattice 
$L'$ of $L$, an integer $n$ and a negative-definite lattice $M$ such that
\[
L'\cong U\oplus E_{8}^{n}\oplus M,
\]
with $\rk M\le 11$ and such that the root part of $M$ does not contain any $E_{8}$ or an ADE lattice of rank at least $9$.}

Using the isometry $U\oplus E_8^3\cong U\oplus \Lambda$ and that the Leech lattice $\Lambda$ has no $(-2)$-root, one deduces that ``the exceptional lattice $E(L)$ is contained in $M$.'' Since there are ``only finitely many possibilities for the root part of $M$,'' the argument given later checks all cases and shows ``that $E(L)=0$ for all of them,'' which ``conclud[es] the proof of the main theorem.''

\end{tcolorbox}
\begin{tcolorbox}[
    enhanced,
    colback=green!4,
    colframe=green!55!black,
    colbacktitle=green!18,
    coltitle=black,
    boxrule=0.8pt,
    arc=2mm,
    left=6pt,
    right=6pt,
    top=6pt,
    bottom=6pt,
    title=\textbf{{\LiveMathematicianBench} Example  \refstepcounter{promptbox}\thepromptbox~ (Question \& Answers)\label{box:universal}},
]
\footnotesize

\textbf{Generated Question} \par\smallskip
Here, a hyperbolic lattice means an even integral lattice of signature $(1,\operatorname{rk} L-1)$, and the exceptional lattice of $L$ is the sublattice of vectors in $L$ with finite orbit under $\mathrm{O}^+(L)/W(L)$, where $\mathrm{O}^+(L)$ is the group of positive isometries of $L$ and $W(L)$ is the subgroup generated by reflections in $(-2)$-roots. Which is the strongest statement that holds for every hyperbolic lattice $L$ of rank at least $46$?

\vspace{1em}
\hrule
\vspace{1em}

\textbf{Answer Choices} \par\smallskip
\textbf{A.} One of the remaining options is correct, but a stronger result can be proven. \textcolor{green!60!black}{\checkmark}

\textbf{B.} For every hyperbolic lattice $L$ of rank at least $46$, the lattice $L$ itself is isometric to $U\oplus E_8^n\oplus M$ for some $n\ge 5$ and some negative-definite lattice $M$ of rank at most $11$ whose root part contains no $E_8$ or ADE summand of rank at least $9$.

\textbf{C.} The exceptional lattice of every hyperbolic lattice $L$ of rank at least $46$ has rank at most $11$.

\textbf{D.} For every hyperbolic lattice $L$ of rank at least $46$, if $L'$ is a maximal overlattice of $L$ with $L'\cong U\oplus E_8^n\oplus M$ and $n\ge 5$, then the exceptional lattice of $L$ is exactly the root sublattice of $M$.

\textbf{E.} For every hyperbolic lattice $L$ of rank at least $46$, after passing to a maximal overlattice $L'\cong U\oplus E_8^n\oplus M$ with $n\ge 5$, the isometry $U\oplus E_8^3\cong U\oplus\Lambda$ with the Leech lattice $\Lambda$ already forces the exceptional lattice of $L$ to be trivial, without any further case-by-case analysis of $M$.

\end{tcolorbox}

\newpage
\paragraph{13. Universal--Existential.}
\emph{Canonical form:} $\forall\, x\; \exists\, y,\; P(x,y)$.
This category covers theorems asserting that for every object in a class, there exists an associated object satisfying a property. The existential witness may depend on the universally quantified variable.
\emph{Example:} ``For every $\varepsilon > 0$, there exists $\delta > 0$ such that $|f(x) - f(a)| < \varepsilon$ whenever $|x - a| < \delta$.''

\begin{tcolorbox}[
    enhanced,
    colback=blue!4,
    colframe=blue!55!black,
    colbacktitle=blue!18,
    coltitle=black,
    boxrule=0.8pt,
    arc=2mm,
    left=6pt,
    right=6pt,
    top=6pt,
    bottom=6pt,
    title=\textbf{{\LiveMathematicianBench} Example (Theorem \& Sketch)},
]
\footnotesize

\textbf{Original Theorem} (\cite{mendonca2025graphsasymmetricramseyproperties}) \par\smallskip
\textbf{Theorem} (Ne{\v{s}}et{\v{r}}il \& R{\"o}dl, 1976)\label{thm_NR}\textbf{.}
For every $k\geq 2$ there is a graph $G$ such that
$K_k \nsubseteq G$ and $G\rightarrow K_{k-1}$.

\vspace{1em}
\hrule
\vspace{1em}

\textbf{Proof Sketch} \par\smallskip
The post-theorem introduction gives the following proof outline for the main theorem: it ``combines probabilistic methods with the hypergraph container framework Balogh--Morris--Samotij, \textit{Independent sets in hypergraphs} (2015) and Saxton--Thomason, \textit{Hypergraph containers} (2015) and is inspired by ideas from Bollob\'as--Erd\H{o}s, \textit{Ramsey graphs} (2001).'' The paper is then organized into steps: (1) Next, ``with high probability \dots the graph $G$ obtained in a natural way from every `dense' subhypergraph of a suitable $n$-vertex random $s$-uniform hypergraph satisfies $G\rightarrow(K_s,K_{k-1})$ for $s=R(k)-1$.'' (2) Then, ``with high probability a suitable random hypergraph $\mathcal{H}$ contains a dense subhypergraph $\mathcal{H}_0$ that will allow us to obtain a graph $G$ such that $G\not\rightarrow K_k$.'' (3) ``These results are then combined later'' to produce a graph $G$ satisfying both $G\not\rightarrow K_k$ and $G\rightarrow(K_s,K_{k-1})$ (with $s=R(k)-1$).

\end{tcolorbox}
\begin{tcolorbox}[
    enhanced,
    colback=green!4,
    colframe=green!55!black,
    colbacktitle=green!18,
    coltitle=black,
    boxrule=0.8pt,
    arc=2mm,
    left=6pt,
    right=6pt,
    top=6pt,
    bottom=6pt,
    title=\textbf{{\LiveMathematicianBench} Example  \refstepcounter{promptbox}\thepromptbox~ (Question \& Answers)\label{box:uniexi}},
]
\footnotesize

\textbf{Generated Question} \par\smallskip
Let $K_t$ denote the complete graph on $t$ vertices. For graphs $G$ and $H$, write $G\rightarrow H$ to mean that every red-blue coloring of the edges of $G$ contains a monochromatic copy of $H$. Which is the strongest statement that holds for every integer $k\ge 2$?

\vspace{1em}
\hrule
\vspace{1em}

\textbf{Answer Choices} \par\smallskip
\textbf{A.} One of the remaining options is correct, but a stronger result can be proven. \textcolor{green!60!black}{\checkmark}

\textbf{B.} For every integer $k\ge 2$ there is a graph $G$ such that $K_k\nsubseteq G$ and $G\rightarrow (K_k,K_{k-1})$.

\textbf{C.} For every integer $k\ge 2$ there is a graph $G$ such that $G\rightarrow K_{k-1}$.

\textbf{D.} For every integer $k\ge 2$ there is a graph $G$ such that $K_k\nsubseteq G$ and $G\rightarrow (K_{R(k)-1},K_{k-1})$, where $R(k)$ denotes the Ramsey number of $K_k$.

\textbf{E.} For every integer $k\ge 2$ there is a graph $G$ such that $G\nrightarrow K_k$ and $G\rightarrow (K_{R(k)-1},K_k)$, where $R(k)$ denotes the Ramsey number of $K_k$.

\end{tcolorbox}

\section{Benchmark Composition Across Construction Stages}
\label{app:benchmark_composition}

\begin{figure}[t]
\centering
\includegraphics[width=0.6\linewidth]{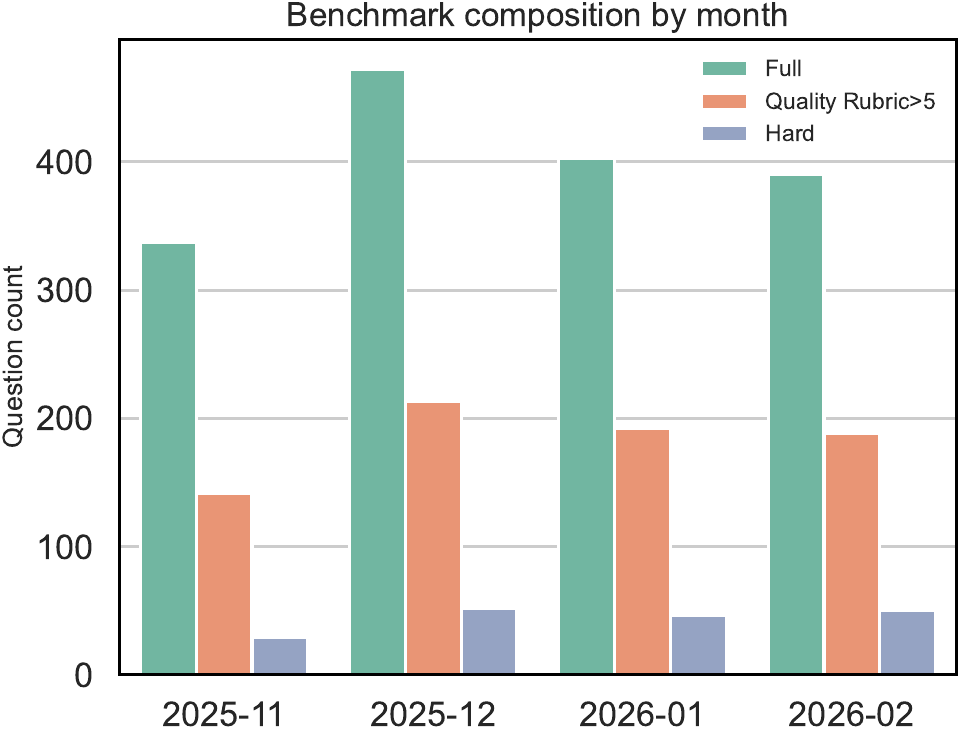}
\caption{Benchmark composition by month across major construction stages. \textbf{Full} is the complete generated candidate pool, \textbf{Quality Rubric$>5$} is the subset passing the rubric threshold, and \textbf{Hard} is the final released split after the remaining filtering and hardness-selection steps.}
\label{fig:benchmark_composition}
\end{figure}

Figure~\ref{fig:benchmark_composition} summarizes how the benchmark size changes across the main construction stages for each month. \textbf{Full} denotes the complete post-generation candidate pool, \textbf{Quality Rubric$>5$} denotes the subset retained after the rubric-based quality filter, and \textbf{Hard} denotes the final released split after stem-nontrivial filtering and hardness calibration. The figure shows that the pipeline consistently reduces the candidate pool in a structured way rather than through ad hoc pruning: each month begins from a substantially larger generated set, is narrowed by quality control, and is then further distilled into a compact hard benchmark. 

\section{Validation of Distractor Generation}
\label{app:distractor_ablation}

To directly evaluate the design choices
used in distractor generation,
we conduct a fixed-theorem ablation
over 98 benchmark items.

Across all conditions,
we hold the source theorem,
question stem,
and correct option fixed,
and vary only three components
of false-option generation:

(i) access to the proof sketch,
(ii) category-aware distractor guidance,
and (iii) the adversarial revision pass.

We report solver accuracy,
where lower accuracy indicates
more challenging distractors.

\begin{table}[t]
\centering
\small
\begin{tabular}{@{}lcccccc@{}}
\toprule
Setting
& Sketch
& Category
& Revision
& Acc.
& Hardness
& $n$ \\
\midrule
Minimal baseline
& No
& No
& No
& 72.4\%
& 27.6\%
& 98 \\

w/o sketch
& No
& Yes
& Yes
& 75.5\%
& 24.5\%
& 98 \\

w/o category guidance
& Yes
& No
& Yes
& 63.3\%
& 36.7\%
& 98 \\

w/o revision
& Yes
& Yes
& No
& 62.2\%
& 37.8\%
& 98 \\

Full pipeline
& Yes
& Yes
& Yes
& \textbf{61.2\%}
& \textbf{38.8\%}
& 98 \\
\bottomrule
\end{tabular}

\caption{
Fixed-theorem ablation of distractor generation.
Sketch denotes access to the proof sketch;
Category denotes category-aware distractor guidance;
and Revision denotes the adversarial revision pass.
Hardness is defined as one minus solver accuracy.
}

\label{tab:distractor_ablation}
\end{table}

The full pipeline produces
the most challenging distractors,
reducing solver accuracy
from 72.4\% under the minimal baseline
to 61.2\%,
an 11.2 percentage-point decrease.

Removing access to the proof sketch
increases accuracy to 75.5\%,
indicating that sketch access
makes the largest contribution
to distractor difficulty.

Removing category-aware guidance
increases accuracy by 2.1 percentage points,
while removing the adversarial revision pass
increases accuracy by 1.0 percentage point.

These results support the use
of proof sketches and category-aware guidance
for constructing more challenging distractors,
while the revision pass
provides a smaller additional contribution.

\section{Evaluation Implementation Details}
\label{app:evaluation}

This section documents the released evaluation implementation in \texttt{LiveMathematicianBench/eval/}. The current repository provides three backend-specific entry points: \texttt{eval.py} for Azure OpenAI models, \texttt{eval\_claude.py} for Claude models accessed through the Anthropic API, and \texttt{eval\_vllm.py} for models served through an OpenAI-compatible vLLM endpoint. Despite backend-specific client code and usage accounting, the three scripts share the same hard-set loader, prompt template, deterministic choice shuffling, answer parser, correctness rule, and output JSON structure.

\paragraph{Benchmark file format.}
For each evaluation month, the scripts load the hard benchmark split from
\verb|data/<month>/hard/qaEval_<month>_ge5_hard.json|.
Each JSON item contains a unique identifier and an \texttt{mcq} object with a question stem, one \texttt{correct\_choice}, four distractors in \texttt{choices}, and auxiliary metadata such as \texttt{mcq.meta.score}. During evaluation, the released scripts consume the finalized \texttt{mcq} fields only; source theorem context, sketches, and other record-level annotations remain in the benchmark file but are not inserted into the released prompts.

\paragraph{Prompt construction.}
Evaluation is implemented as a five-way multiple-choice selection task. For each item, the scripts combine the stored correct choice with the four distractors, then apply a deterministic shuffle using a per-item seed equal to the global seed plus the item's dataset index. The shuffled options are relabeled as \texttt{A}--\texttt{E}. The model then receives a fixed system prompt instructing it to act as an expert mathematician, reason step by step, and place its final answer inside \verb|\boxed{}|. The user prompt contains only two blocks:
\begin{enumerate}
    \item the question stem from \texttt{mcq.question}; and
    \item the five relabeled answer choices.
\end{enumerate}
The choices are formatted as ``\texttt{(A) ...}'', ``\texttt{(B) ...}'', and so on, separated by blank lines. All models were evaluated in the without-sketch setting. Due to budget constraints, we additionally ran sketch-aware evaluation only for the few models that performed best in the without-sketch setting.

\paragraph{Answer extraction and scoring.}
After generation, the scripts recover the model's predicted label by first searching for the last occurrence of \verb|\boxed{A}| through \verb|\boxed{E}|. If no boxed answer is found, the parser falls back to either a one-character response in \texttt{A}--\texttt{E} or, failing that, the last standalone capital letter in \texttt{A}--\texttt{E}. In the Azure and vLLM evaluators, answer extraction is applied directly to the returned response text. In the Claude evaluator, the Anthropic response is split into text and thinking blocks; the final prediction is extracted from the text block, while the thinking block is recorded separately. An item is marked correct iff the extracted label matches the post-shuffle label of the ground-truth option. Monthly accuracy is then computed as \texttt{correct/total} over the evaluated hard split.

\paragraph{Execution details.}
All evaluators parallelize item-level inference with a \texttt{ThreadPoolExecutor} controlled by \texttt{--concurrency}. They also support repeated sampling through \texttt{--n}; in that case each question is answered multiple times, a full per-sample list is stored under \texttt{samples}, and the top-level prediction fields intentionally mirror the first sample for backward compatibility. For every sample, the scripts record elapsed time, prompt tokens, completion tokens, total tokens, and reasoning-token counts whenever the backend exposes them.

The backend-specific command-line options differ as follows. The Azure evaluator requires \texttt{--endpoint} and \texttt{--api-key}, accepts \texttt{--api-version}, exposes both a client-wide \texttt{--timeout} and a per-sample \texttt{--request-timeout}, and optionally switches from chat completions to the Responses API through \texttt{--use-responses-api}; this switch is also enabled automatically when the model name contains \texttt{gpt-5.4}. The Claude evaluator requires \texttt{--base-url}, accepts \texttt{--api-key}, supports \texttt{--thinking-budget} to request extended thinking, and exposes \texttt{--debug}, \texttt{--timeout}, and \texttt{--request-timeout}. The vLLM evaluator requires \texttt{--base-url}, accepts an optional \texttt{--api-key}, exposes \texttt{--temperature}, \texttt{--top-p}, \texttt{--debug}, \texttt{--timeout}, and \texttt{--request-timeout}, and also supports optional \texttt{--use-responses-api} with automatic activation for model names containing \texttt{gpt-5.4}.

Token accounting is normalized across backends. Azure records usage directly from the provider response. Claude and vLLM additionally normalize \texttt{completion\_tokens} so that, when reasoning-token metadata is available, the reported completion count includes both visible output tokens and reasoning tokens, matching the OpenAI-style accounting used elsewhere in the benchmark.

\paragraph{Resuming and result serialization.}
All scripts support fault-tolerant resumption through \texttt{--resume}. If an output file already exists, the evaluator loads the previous JSON, skips items whose prior record has a non-\texttt{None} \texttt{model\_answer} and no recorded error, and re-runs unanswered or failed items only. Results are written to
\verb|results/<month>/accuracy_test_<model>_<month>_<effort>.json|,
where model names are sanitized for filesystem safety and \texttt{<effort>} defaults to \texttt{default} when no reasoning level is specified. Each output file contains (i) \texttt{test\_info} metadata such as model name, month, seed, maximum token budget, number of samples, and timestamps; (ii) duplicated aggregate accuracy summaries under \texttt{summary.all} and \texttt{overall}; and (iii) \texttt{detailed\_results} records containing the item identifier, extracted \texttt{model\_answer}, shuffled \texttt{correct\_answer}, \texttt{raw\_response}, correctness flag, latency, token usage, \texttt{reasoning\_tokens}, reasoning effort, \texttt{n\_samples}, optional per-sample \texttt{samples}, auxiliary \texttt{mcq.meta.score}, and any backend error message. For Claude runs, each record additionally stores \texttt{raw\_thinking} extracted from the Anthropic thinking block.

\section{Additional Evaluation Results}
\subsection{Sketch-Aware Accuracy with Substitution-Resistant Settings}
\label{app:sketch_choice_style}
\begin{figure}[t]
\centering
\includegraphics[width=0.6\linewidth]{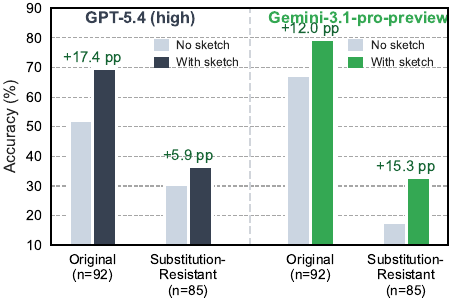}
\caption{Sketch-aware accuracy gains broken down by choice style. For each model, we compare accuracy without sketch access and with sketch access on the original-choice subset and the substitution-resistant subset. Proof sketches improve accuracy in both regimes, but the relative gain differs across models and is particularly pronounced on the harder substitution-resistant items for Gemini-3.1-pro-preview.}
\label{fig:sketch_choice_style}
\end{figure}

Built upon the sketch-aware evaluation results in Figure~\ref{fig:choice_style_accuracy}(b), we further analyze how the presence of proof sketches interacts with and without substitution-resistant question formulations. We break down sketch-aware gains, separating standard multiple-choice questions from substitution-resistant items. Figure~\ref{fig:sketch_choice_style} shows that proof sketches improve performance in both settings, but the magnitude of the gain depends on the model and the choice style. GPT-5.4 benefits more strongly on the original-choice subset, whereas Gemini-3.1-pro-preview shows especially large gains on the substitution-resistant subset. This suggests that proof-level guidance can help models not only identify the correct theorem statement, but also reason through adversarial answer formulations designed to resist shortcut strategies.

\subsection{Cost Analysis} 
\begin{figure*}
\centering
\includegraphics[width=0.6\linewidth]{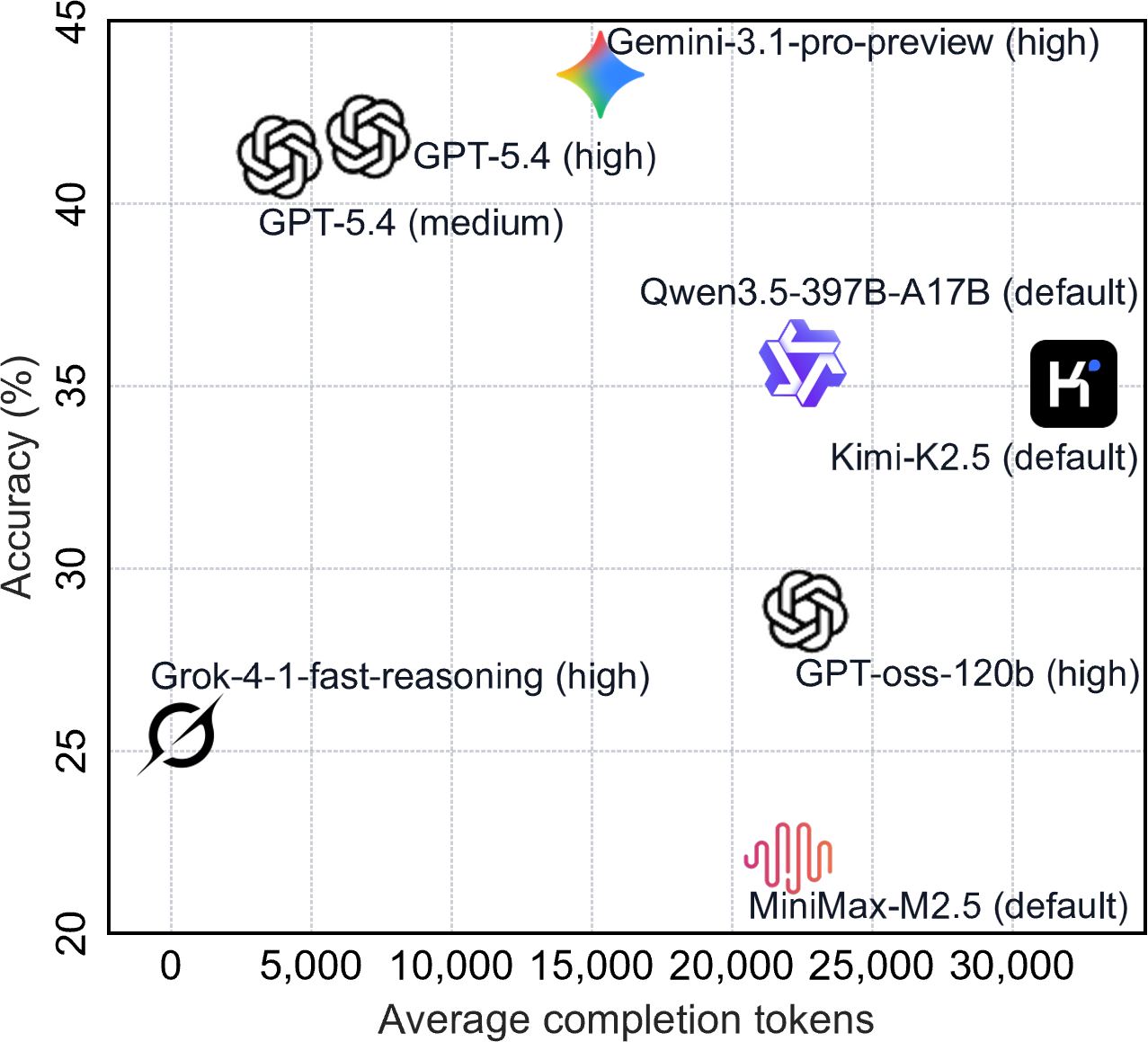}
\caption{Accuracy versus average completion tokens on \LiveMathematicianBench. Each point denotes one model setting. The most favorable region is the upper-left, which corresponds to higher accuracy with fewer generated tokens. GPT-5.4 lies on a particularly strong frontier: it remains close to Gemini-3.1-pro-preview in accuracy while using substantially fewer completion tokens.}
\label{fig:efficiency_frontier}
\end{figure*}

Accuracy alone does not capture how practical a model is for benchmark-scale evaluation. Figure~\ref{fig:efficiency_frontier} shows that the accuracy-cost frontier is led by GPT-5.4 rather than by the highest-token models. Gemini-3.1-pro-preview achieves the top accuracy at 43.5\%, but it requires about 15.3k average completion tokens. By contrast, GPT-5.4 (high) reaches a similar 41.8\% accuracy with only about 7.0k tokens, and GPT-5.4 (medium) remains close at 41.2\% with about 3.8k tokens. In other words, GPT-5.4 combines high accuracy with materially lower token cost, while several other long-generation models, such as Qwen and Kimi, use far more tokens without matching GPT-5.4 or Gemini in accuracy.

At the same time, several open-weight or alternative frontier systems occupy distinct efficiency niches. Qwen3.5-397B-A17B and Kimi-K2.5 achieve mid-tier accuracy at higher token budgets, while GPT-oss-120b generates even more tokens without matching the top closed-model accuracy band. This pattern suggests that improvements in research-level mathematical reasoning cannot be reduced to simply allocating more inference budget: architectural and training differences still materially affect how efficiently models convert tokens into correct mathematical judgments.

\section{Prompt Details}\label{app:prompt}
We provide below some examples of system prompts used in our construction pipeline. We will focus on prompts used in Stage 4 (Question-Answer Pair Generation) as it best demonstrates the methodological novelty of \LiveMathematicianBench. 

To show case our \emph{category-specific} system prompts, we include prompts from two categories: \emph{biconditional/equivalence} and \emph{implication} as they are representative of our central design philosophy.
\paragraph{Stage 4a: Question Stem and Correct Choice Generation.}
\noindent
\begin{tcolorbox}[
    enhanced,
    colback=purple!4,
    colframe=purple!55!black,
    colbacktitle=purple!18,
    coltitle=black,
    boxrule=0.8pt,
    arc=2mm,
    left=6pt,
    right=6pt,
    top=6pt,
    bottom=6pt,
    title=\textbf{System Prompt for Biconditional/Equivalence Theorems},
]
\footnotesize

You are an expert in mathematics. You are given one input:
\begin{itemize}
    \item \texttt{"main\_theorem"}: the full \LaTeX{} code of a theorem-like environment.
\end{itemize}

The \texttt{"main\_theorem"} is an equivalence-type theorem: statements asserting ``A if and only if B'' or ``the following are equivalent.''

\textbf{Task:} Generate a multiple-choice question stem and the single correct answer option that exactly and faithfully represent the full mathematical content of the theorem.

\begin{enumerate}
    \item \textbf{Detect the equivalence structure:} 
    Read the theorem and identify any ``if and only if'' or ``the following are equivalent'' pattern.
    List each distinct statement or property that appears in the equivalence.
    Rewrite them as short, self-contained statements such as A, B, C, etc.

    \item \textbf{Decide which side(s) will appear in the question.}
    If there are two statements ($A \Leftrightarrow B$), design the question as ``Which of the following statements is equivalent to A?''. i.e.\ A appears in the stem and B becomes the correct answer. 
    If there are three or more equivalent statements ($A \Leftrightarrow B \Leftrightarrow C \dots$), focus on identifying any new side of the equivalence introduced by the theorem, and write the question as ``Which of the following statements is equivalent to both A and B?''.

    \item \textbf{Write the question stem.}
    Use only the side(s) named in Step 2. Do not include the correct equivalent statement in the stem. Phrase the question clearly, using formal but concise mathematical language. Avoid referencing the theorem itself (``according to the theorem'', etc.).

    Do not use phrases such as:
    \begin{itemize}
        \item ``according to the theorem/statement''
        \item ``the theorem shows''
        \item ``as proved above''
    \end{itemize}

    \item \textbf{Determine the correctness criterion for the correct option.}
    The correct option must satisfy all of the following:
    \begin{itemize}
        \item It captures the full conclusion of the theorem, up to harmless rephrasing;
        \item It includes all essential aspects of the result (structure, existence, bounds, conditional clauses, arithmetic restrictions, etc.);
        \item It does not omit essential components, strengthen or weaken the statement, change quantifiers or scope, or add auxiliary conclusions not present in the theorem.
    \end{itemize}

    \item \textbf{Output requirements.}
    \begin{itemize}
        \item Do NOT generate distractors.
        \item Do NOT mention any proof or proof sketch.
        \item Produce a JSON object with exactly the following fields:
    \end{itemize}
\end{enumerate}

\begin{verbatim}
{
  "question": "<question stem written from main_theorem>",
  "correct_choice": {"label": "A", "text": "<the correct option based on main_theorem>"}
}
\end{verbatim}

\end{tcolorbox}

\noindent
\begin{tcolorbox}[
    enhanced,
    colback=purple!4,
    colframe=purple!55!black,
    colbacktitle=purple!18,
    coltitle=black,
    boxrule=0.8pt,
    arc=2mm,
    left=6pt,
    right=6pt,
    top=6pt,
    bottom=6pt,
    title=\textbf{System Prompt for Implication Theorems},
]
\footnotesize

You are an expert in mathematics. You are given one input:
\begin{itemize}
    \item \texttt{"main\_theorem"}: the full \LaTeX{} code of a theorem-like environment.
\end{itemize}

The \texttt{"main\_theorem"} is an implication-type theorem, i.e.\ it asserts that certain assumptions imply a specific conclusion (for example: ``If A holds, then B holds'').

\textbf{Task:} Generate a multiple-choice question stem and the single correct answer option that exactly and faithfully represent the full mathematical content of the theorem.

\textbf{Instruction:}
\begin{enumerate}
    \item \textbf{Detect the implication structure:}\\
    Read the theorem and identify:
    \begin{itemize}
        \item the full set of assumptions (A), including parameter ranges, structural hypotheses, and problem setup;
        \item the conclusion (B).
    \end{itemize}
    Rewrite these internally as self-contained statements, but do not simplify or strengthen them.

    \item \textbf{Write the question stem}\\
    The question stem must:
    \begin{itemize}
        \item explicitly include all essential assumptions of the theorem (including equations, boundary conditions, parameter ranges, and structural hypotheses);
        \item ask the solver to identify the strongest statement that can be proved under these assumptions;
        \item avoid referencing the theorem itself (``according to the theorem'', etc.).
    \end{itemize}

    Do not use phrases such as:
    \begin{itemize}
        \item ``according to the theorem/statement''
        \item ``the theorem shows''
        \item ``as proved above''
    \end{itemize}

    Acceptable stem formulations include:
    \begin{itemize}
        \item ``What is the strongest statement that can be proved about \dots?''
        \item ``Which of the following is the strongest consequence of the assumptions above?''
    \end{itemize}
    The stem should be general enough that weaker consequences, stronger but unjustified conclusions, and conclusions about closely related properties all appear plausible.

    \item \textbf{Determine the correctness criterion for the correct option}\\
    The correct option must satisfy all of the following:
    \begin{itemize}
        \item It captures the full conclusion of the theorem, up to harmless rephrasing;
        \item It includes all essential aspects of the result (structure, existence, bounds, conditional clauses, arithmetic restrictions, etc.);
        \item It does not omit essential components, strengthen or weaken the statement, change quantifiers or scope, or add auxiliary conclusions not present in the theorem.
    \end{itemize}

    \item \textbf{Output requirements}
    \begin{itemize}
        \item Do NOT generate distractors.
        \item Do NOT mention any proof or proof sketch.
        \item Produce a JSON object with exactly the following fields:
    \end{itemize}
\end{enumerate}

\begin{verbatim}
{
  "question": "<question stem written from main_theorem>",
  "correct_choice": {"label": "A", "text": "<the correct option based on main_theorem>"}
}
\end{verbatim}

\end{tcolorbox}

\newpage
\paragraph{Stage 4b: Sketch-Adversarial Distractor Generation.}
\noindent
\begin{tcolorbox}[
    enhanced,
    breakable,
    colback=purple!4,
    colframe=purple!55!black,
    colbacktitle=purple!18,
    coltitle=black,
    boxrule=0.8pt,
    arc=2mm,
    left=6pt,
    right=6pt,
    top=6pt,
    bottom=6pt,
    title=\textbf{System Prompt for Biconditional/Equivalence Type Theorems},
]
\footnotesize

You are a senior research mathematician generating hard-negative distractors. You are given four inputs:
\begin{itemize}
    \item \texttt{"main\_theorem"}: full \LaTeX{} theorem-like environment.
    \item \texttt{"proof\_sketch"}: proof outline for distractor design only.
    \item \texttt{"question\_stem"}: the question stem (string).
    \item \texttt{"correct\_choice"}: the correct option (string).
\end{itemize}

\textbf{Task:}\\
Generate exactly four choices B--E:
\begin{itemize}
    \item B, D, E: false but highly plausible (hard negatives).
    \item C: weaker-but-true: $A \Rightarrow C$ and $C \nRightarrow A$.
\end{itemize}
Do not edit \texttt{question\_stem} or \texttt{correct\_choice.text}.

\textbf{Mandatory internal workflow} (do internally; then output metadata):
\begin{itemize}
    \item \textbf{Read \& parse everything} (including the sketch): fully read \texttt{main\_theorem}, \texttt{proof\_sketch}, \texttt{question\_stem}, and Option A; identify central objects, hypotheses, quantifiers, case splits, and what the sketch highlights as ``hard steps.''
    \item \textbf{Extract sketch load-bearers:} from the sketch, identify 1--3 load-bearing steps and the constraints they enforce (ranges, excluded cases, dependence of constants, uniform vs non-uniform, effective vs non-effective, quantifier order, hidden hypotheses, structural restrictions).
    \item \textbf{Design distractors via sketch-derived constraint tampering:}
    \begin{itemize}
        \item Each false distractor must violate a specific sketch-derived constraint and be really difficult to rule out without the sketch.
        \item The weaker-true distractor C must be implied by A by dropping one critical component.
    \end{itemize}
    \item \textbf{Keyword normalization:} spread $\ge 2$ key technical terms from A across $\ge 2$ of B/D/E.
    \item \textbf{Near-miss requirement:} every distractor must differ from A in only one or two mathematically meaningful places. Avoid obviously unrelated statements, blatant negations, or options that can be rejected by surface reading alone.
    \item \textbf{Anti-triviality requirement:} make the solver distinguish options by subtle issues such as quantifier order, dependence of constants, exact asymptotic regime, necessity of an extra hypothesis, sharpness of the bound, uniqueness vs mere existence, completeness of a classification, or uniformity vs pointwise validity.
\end{itemize}

\textbf{False distractors: required breakdown}
\begin{itemize}
    \item Produce three false options (B, D, E) with:
    \begin{itemize}
        \item Two template-based (choose two different): boundary/range trap; uniformity/effectivity mirage; quantifier/dependence trap; property confusion; stronger-theorem trap.
        \item One wildcard false distractor (exactly one of B/D/E): does not need to match a template, but must still be tightly coupled to a specific sketch-derived constraint.
        \item record its label in \texttt{meta.wildcard\_false\_label}.
    \end{itemize}
\end{itemize}

\textbf{Surface anti-gaming constraints:}
\begin{itemize}
    \item B, D, E: match A in logical shape (quantifiers/clauses), \LaTeX{}/technical density, and approximate length.
    \item C may be slightly shorter (natural weakening).
    \item No hedging unless A uses it. Proper \LaTeX{} syntax.
    \item Do not make any distractor obviously false from a missing definition, gross type mismatch, or dramatic change of topic.
    \item If A contains several clauses, each false distractor should tamper with only one load-bearing clause, keeping the rest close to A.
    \item At least two of B/D/E should look locally compatible with the hypotheses and fail only on a subtle point that an expert would have to inspect carefully.
\end{itemize}

\textbf{Output format:}\\
Return a single JSON object, for example:
\begin{verbatim}
{
  "choices": [
    {"label": "B", "text": "<false distractor>"},
    {"label": "C", "text": "<weaker-true distractor>"},
    {"label": "D", "text": "<false distractor>"},
    {"label": "E", "text": "<false distractor>"}
  ],
  "meta": {
    "weaker_true_label": "C",
    "false_labels": ["B", "D", "E"],
    "wildcard_false_label": "<one of B/D/E>"
  },
  "sketch_usage_meta": [
    {
      "label": "B",
      "sketch_hook_type": "<one of: case_split | counting_estimate | trace_identity | 
                          geometric_construction | computational_check | 
                          outer_automorphism | finiteness | characteristic | 
                          regularity | other>",
      "tampered_component": "<short identifier of the sketch-derived constraint 
      you tampered with>",
      "template_used": "<one of: boundary_range | uniformity_effectivity | 
                       quantifier_dependence | property_confusion | 
                       stronger_trap | wildcard>"
    },
    {
      "label": "C",
      "sketch_hook_type": "<same enum>",
      "tampered_component": "<what critical component of A was dropped to 
      make C weaker-true>",
      "template_used": "weaker_true"
    },
    {
      "label": "D",
      "sketch_hook_type": "<same enum>",
      "tampered_component": "<short identifier>",
      "template_used": "<one of: boundary_range | uniformity_effectivity | 
                       quantifier_dependence | property_confusion | 
                       stronger_trap | wildcard>"
    },
    {
      "label": "E",
      "sketch_hook_type": "<same enum>",
      "tampered_component": "<short identifier>",
      "template_used": "<one of: boundary_range | uniformity_effectivity | 
                       quantifier_dependence | property_confusion | 
                       stronger_trap | wildcard>"
    }
  ]
}
\end{verbatim}

\end{tcolorbox}

\end{document}